\documentclass[letterpaper]{article} % DO NOT CHANGE THIS
\usepackage{aaai2026}  % DO NOT CHANGE THIS
\usepackage{times}  % DO NOT CHANGE THIS
\usepackage{helvet}  % DO NOT CHANGE THIS
\usepackage{courier}  % DO NOT CHANGE THIS
\usepackage[hyphens]{url}  % DO NOT CHANGE THIS
\usepackage{graphicx} % DO NOT CHANGE THIS
\usepackage{natbib}  % DO NOT CHANGE THIS AND DO NOT ADD ANY OPTIONS TO IT
\usepackage{caption} % DO NOT CHANGE THIS AND DO NOT ADD ANY OPTIONS TO IT
\usepackage{subcaption}
\usepackage{booktabs}
\usepackage{multirow}
\usepackage{makecell}
\usepackage{amssymb}
\usepackage{amsmath}

\usepackage{algorithm}
\usepackage{algorithmic}

\usepackage{newfloat}
\usepackage{listings}
\DeclareCaptionStyle{ruled}{labelfont=normalfont,labelsep=colon,strut=off} % DO NOT CHANGE THIS
\floatstyle{ruled}
\newfloat{listing}{tb}{lst}{}
\floatname{listing}{Listing}
\title{JoDiffusion: Jointly Diffusing Image with \\ Pixel-Level Annotations for Semantic Segmentation Promotion}
\author {
    Haoyu Wang\textsuperscript{\rm 1},
    Lei Zhang\textsuperscript{\rm 1}\footnotemark[2],
    Wenrui Liu\textsuperscript{\rm 1},
    Dengyang Jiang\textsuperscript{\rm 1},
    Wei Wei\textsuperscript{\rm 1},
    Chen Ding\textsuperscript{\rm 2}
}
\affiliations {
    \textsuperscript{\rm 1}School of Computer Science, Northwestern Polytechnical University \\
    \textsuperscript{\rm 2}School of Computer Science \& Technology, Xi'an University of Posts \& Telecommunications\\
    wanghaoyunwpu@mail.nwpu.edu.cn, nwpuzhanglei@nwpu.edu.cn
}

\usepackage{bibentry}
\begin{document}

\maketitle
\renewcommand{\thefootnote}{\fnsymbol{footnote}}
\footnotetext[2]{Corresponding author.}
\footnotetext[1]{Code is available at https://github.com/00why00/JoDiffusion.}

\begin{abstract}
Given the inherently costly and time-intensive nature of pixel-level annotation, the generation of synthetic datasets comprising sufficiently diverse synthetic images paired with ground-truth pixel-level annotations has garnered increasing attention recently for training high-performance semantic segmentation models. However, existing methods necessitate to either predict pseudo annotations after image generation or generate images conditioned on manual annotation masks, which incurs image-annotation semantic inconsistency or scalability problem. To migrate both problems with one stone, we present a novel dataset generative diffusion framework for semantic segmentation, termed JoDiffusion. Firstly, given a standard latent diffusion model, JoDiffusion incorporates an independent annotation variational auto-encoder (VAE) network to map annotation masks into the latent space shared by images. Then, the diffusion model is tailored to capture the joint distribution of each image and its annotation mask conditioned on a text prompt. By doing these, JoDiffusion enables simultaneously generating paired images and semantically consistent annotation masks solely conditioned on text prompts, thereby demonstrating superior scalability. Additionally, a mask optimization strategy is developed to mitigate the annotation noise produced during generation. Experiments on Pascal VOC, COCO, and ADE20K datasets show that the annotated dataset generated by JoDiffusion yields substantial performance improvements in semantic segmentation compared to existing methods.
\end{abstract}

\section{Introduction}
\label{sec:intro}

Semantic segmentation plays a crucial role %is one of the core tasks of 
in computer vision, which aims to assign a semantic label to each pixel. It has %key applications
shown promising potential in plenty of practical applications including autonomous driving~\cite{driving}, medical image analysis~\cite{medical} and robot navigation~\cite{navigation} etc.. Although deep neural networks %learning models 
have made significant progress in this task~\cite{semantic}, their pleasing performance highly depends on a high-quality training dataset comprising large-scale paired images and ground-truth pixel-level annotations. %annotated data. 
However, due to high spatial resolution and diverse visual content, %image data acquisition and 
pixel-level manual annotation on image data is prohibitively costly and time-consuming, particularly in complex scenarios characterized by multi-object interaction or dense small-object distribution. %, the difficulty of manual annotation is further exacerbated. 
This bottleneck significantly limits the adoption and deployment of semantic segmentation networks in real scenarios.

\begin{figure}[tbp]
  \centering
  \begin{subfigure}{\linewidth}
    \centering
    \includegraphics[width=1.0\linewidth]{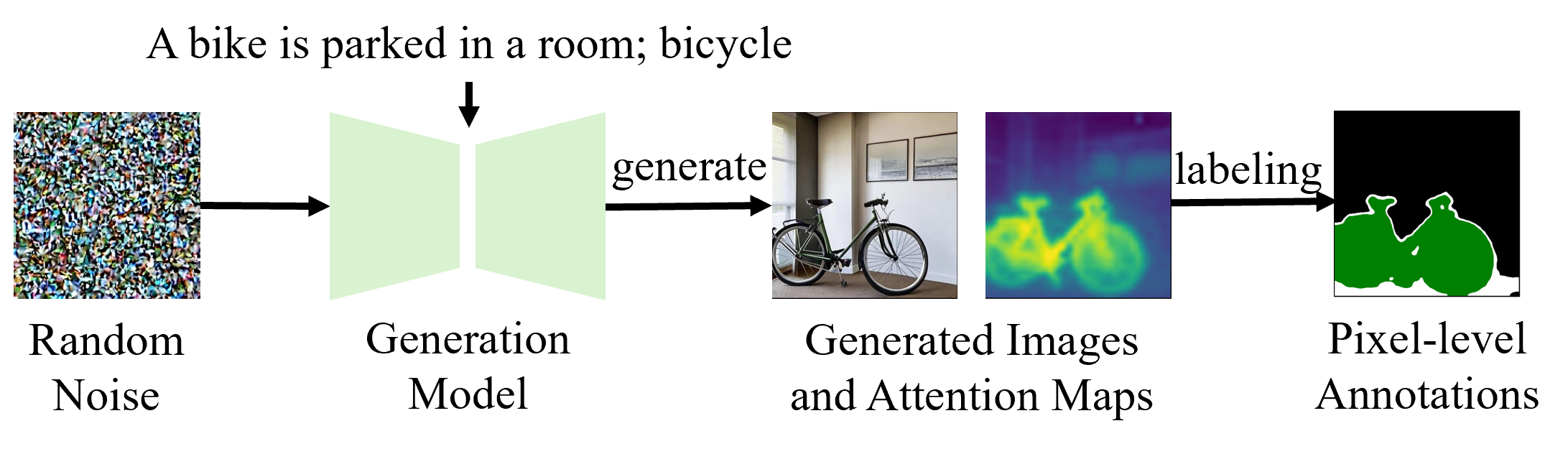}
    \caption{Image2Mask pipeline.}
    \label{fig:1a}
  \end{subfigure}
  \hfill
  \begin{subfigure}{\linewidth}
    \centering
    \includegraphics[width=0.6\linewidth]{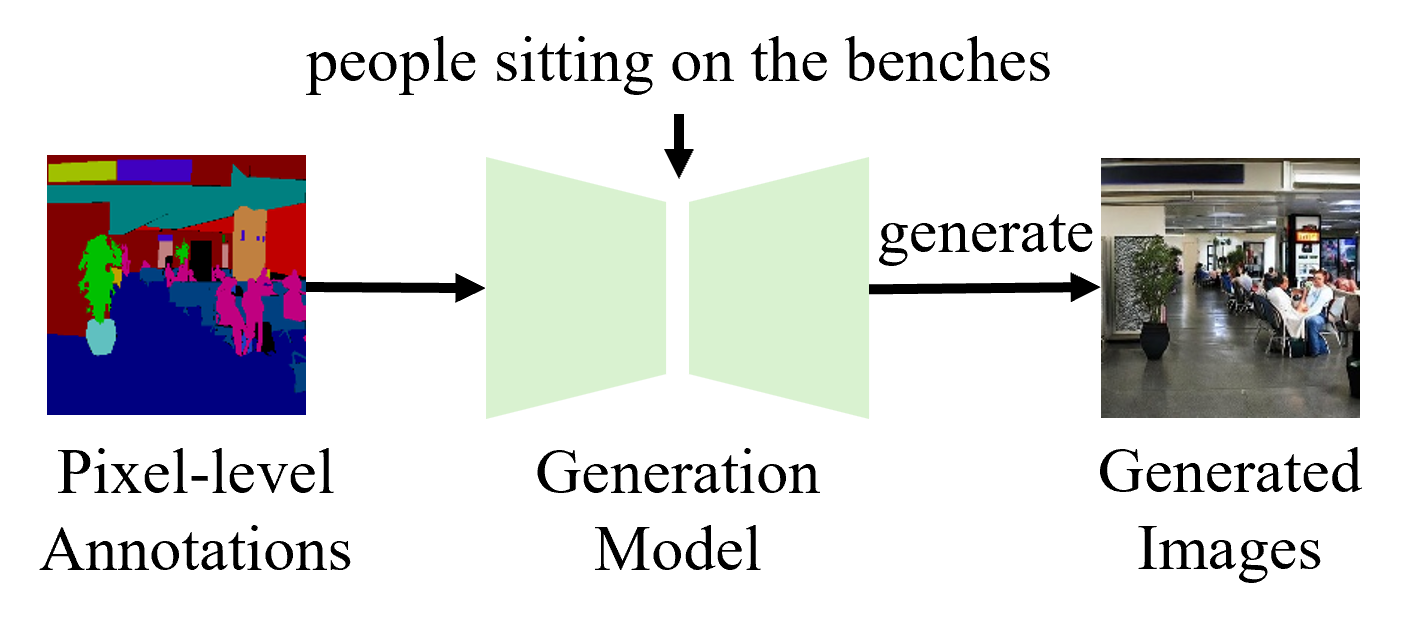}
    \caption{Mask2Image pipeline.}
    \label{fig:1b}
  \end{subfigure}
  \begin{subfigure}{\linewidth}
    \centering
    \includegraphics[width=0.8\linewidth]{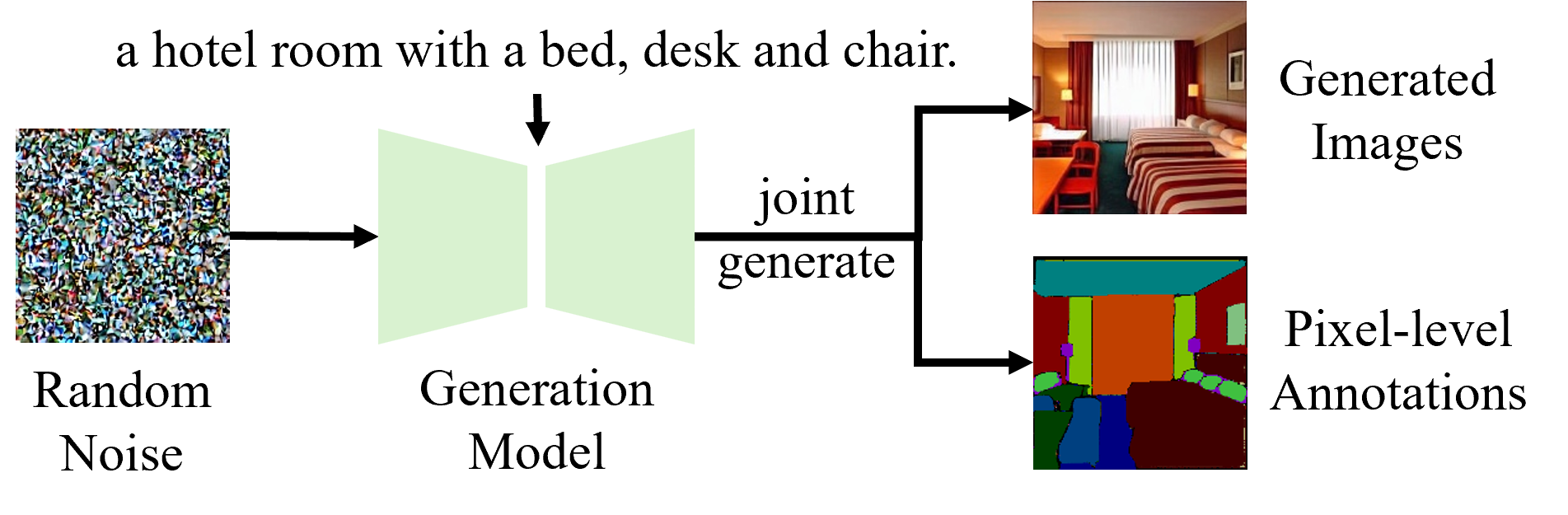}
    \caption{JoDiffusion pipeline.}
    \label{fig:1c}
  \end{subfigure}
  \caption{Comparison of the proposed method with Image2Mask and Mask2Image pipelines. Compared with the other two step-by-step methods, JoDiffusion can directly generate images and corresponding pixel-level annotations.}
  \label{fig:motivation}
\end{figure}

%To reduce the cost of data collection, some works focus on training data generation, trying to generate synthetic samples through generative models to replace or expand real data. 
Inspired by the great success of deep generative models in image synthesis~\cite{sd,sdxl,sd3}, a promising solution lies in generating a synthetic dataset comprising sufficiently diverse synthetic images paired with ground-truth pixel-level annotations. Different from image generation for classification task~\cite{classification}, the dataset generation for semantic segmentation involves generation for paired image and pixel-level annotations. To this end, two lines of research have been investigated, including the Image2Mask~\cite{diffumask,datasetdiffusion,sds} and Mask2Image~\cite{freemask,seggen}.   
%Previous works~\cite{datasetgan,bigdatasetgan} extracted semantic information from the latent space of GAN~\cite{gan}, generate images and extracted pixel-level annotations from features through an additional decoder. With the excellent performance of diffusion models~\cite{ddpm,ddim} in high-quality image generation, more and more works begin to explore how to use diffusion models for data generation. Although it has been widely used in fields such as image classification~\cite{classification} and large language model~\cite{language}, for semantic segmentation tasks, the key challenge is to simultaneously generate high quality images and their corresponding pixel-level segmentation masks.
%Existing semantic segmentation dataset generation methods based on diffusion models can be divided into two types: Image2Mask and Mask2Image. The pipeline of the Image2Mask method~\cite{diffumask,datasetdiffusion,sds} is 
As shown in Fig.~\ref{fig:1a}, the Image2Mask framework initially employs a text-to-image diffusion model to generate synthetic images, and then a cross-attention-based pseudo-annotation scheme is applied to predict pixel-level pseudo annotations by leveraging text-image similarity computed in a latent feature space. %extracts category-related salient areas from the cross-attention map between the text prompt and the generated image, and uses post-processing methods to generate pixel-level annotations.
Although this framework enables direct generation of synthetic semantic segmentation datasets conditioned solely on text prompts, the quality of pixel-level pseudo annotations remains suboptimal. Specifically, semantic inconsistencies between generated images and predicted pseudo annotations arise due to inevitable text-image similarity calculation errors and limited spatial resolution of feature maps compared with original image. Training models on such datasets impose ambiguous semantic information, ultimately leading to suboptimal generalization performance during inference. %type of method can generate masks without relying on additional supervision, due to the resolution limitation of the attention map, the generated semantic results are not accurate enough, especially in small objects and complex background scenes, which are prone to semantic confusion, resulting in noise in the generated dataset that affects the performance of downstream tasks. 
In contrast, the Mask2Image framework employs a dedicated diffusion model to generate synthetic images conditioned on both manual pixel-level annotation masks and text prompts, as shown in Fig.~\ref{fig:1b}. While the introduced high-quality annotation masks ensure semantic consistency with the generated images, %introduce the pixel-level manual annotation mask  the semantic mask as condition to guide diffusion model generate real images that meet the target semantics. However, 
the limited availability of manual annotations inherently restricts image content diversity beyond the scope of provided masks, resulting in suboptimal scalability.%these line of most methods use masks in the original dataset for generation, which limits the diversity of generated images. Although SegGen~\cite{seggen} generates diverse mask data through the additionally trained Text2Mask model, the image generation process of this type of method is highly dependent on the conditional control ability of the diffusion model, which leads to a decrease in the matching degree between the pixel-level annotations and the generated image when dealing with dense objects, complex backgrounds or long-tail categories, ultimately affecting the performance of the trained segmentation model.
%
%In order to solve the semantic inconsistency problem caused by the step-by-step generation of images and masks, 

To mitigate both limitations of existing methods, we present a novel semantic segmentation dataset generation framework, termed %this paper proposes 
JoDiffusion. %an image and pixel-level annotations joint generation framework based on the diffusion model. 
As illustrated in Fig.~\ref{fig:1c}, JoDiffusion differs fundamentally from existing frameworks by enabling simultaneous generation of paired images and pixel-level annotation masks through a joint diffusion model conditioned solely on text prompts. This framework not only guarantees semantic consistency between generated images and annotation masks but also achieves good scalability. %unlike the strategies of Image2Mask and Mask2Image pipelines that generate images or pixel-level annotations first and then process another, JoDiffusion directly models the joint distribution of them, so that the diffusion model can simultaneously generate high-quality images and pixel-accurate semantic masks during the generation process. 
To achieve this goal, we first establish a baseline framework leveraging a standard latent text-image diffusion model and integrate an annotation-specific variational auto-encoder (VAE) network to model the latent distribution of pixel-level annotations. This architecture enables paired images and pixel-level annotation masks to be mapped into a unified latent space, thereby facilitating the maintenance of semantic consistency during the generation process.
% Since semantic masks have less texture and detail information than natural images, lightweight VAE can reduce the dimension of the latent space, making the diffusion model more stable when jointly modeling.
% and reducing the noise interference that may be generated when sampling in the high-dimensional latent space, thereby improving the matching degree between the semantic mask and the image.
Then, %use U-ViT~\cite{uvit,unidiffuser} as the basic diffusion model 
the diffusion model is tailored to jointly diffuse and denoise the input text prompts, images, and pixel-level annotation masks in the latent space. More importantly, the text prompts with random noise is forced to jointly recover the latent representation of each paired image and annotation mask during training. By doing these, the diffusion model can capture the joint distribution of paired images and annotation masks. This enables the simultaneous generation of semantically consistent paired images and annotation masks, relying solely on text prompts. Moreover, during the inference phase, without the requirement of additional manual annotation masks as the Mask2Image framework, the diffusion model can flexibly generalize beyond the limited set of manually annotated masks. In addition, we further develop a mask optimization strategy to mitigate the inevitable annotation noise produced during generation. With the generated high-quality synthetic dataset, we can train an effective segmentation model with better generalization performance. To testify this, we evaluate JoDiffusion onto three benchmark datasets including Pascal VOC~\cite{voc}, MS COCO~\cite{coco}, and ADE20K~\cite{ade20k}. The experimental results demonstrate that, compared to several state-of-the-art competitors, training the same semantic segmentation model with the synthetic dataset generated by JoDiffusion leads to substantially better generalization performance.   %so that they are co-generated in the same latent representation space to ensure the structural and semantic consistency of the final generated results. However, scattered erroneous pixel-level annotations are generated since the diffusion model cannot completely remove noise during the generation process. To this end, we further designed an mask optimization strategy. Specifically, for semantic mask with smaller areas, we analyze the mode value of its edge pixels and use it to adjust the category of the target area, thereby enhancing the consistency of the semantic mask and reducing the impact of noise. Finally, we use the generated image and pixel-level annotation pairs to train the semantic segmentation network. Experimental results on the Pascal VOC~\cite{voc}, MS COCO~\cite{coco}, and ADE20K~\cite{ade20k} datasets show that JoDiffusion achieves the most advanced performance compared to the state-of-the-art methods.

In summary, the primary contributions of this work can be succinctly articulated as follows:

\begin{itemize}
    %\item We propose an novel an annotation VAE network to encode pixel-level annotations, which is 6$ \times $ fewer parameters than image VAE, but the reconstruction performance on the validation set of each dataset can reach more than 98\% mIoU.
    \item we propose a novel synthetic dataset generation framework for semantic segmentation. To the best of our knowledge, this is the first attempt to achieve simultaneous generation of semantically consistent paired images and pixel-level annotation masks conditioned solely on text promts.% the JoDiffusion framework, which realizes the end-to-end joint generation of images with pixel-level annotations.
    %, breaking through the limitations of traditional Image2Mask and Mask2Image methods in terms of generation consistency and diversity.
    \item We also develop a mask optimization strategy to effectively mitigate the annotation noise produced during generation.% uses the edge mode to adjust the category of small target areas to improve the quality of generated semantic masks.
    \item We achieve new SOTA semantic segmentation performance when training the model using the generated synthetic dataset. %Experimental results on multiple datasets show that compared with other SOTA methods, using data generated by JoDiffusion for training can improve the performance of the semantic segmentation model, proving the effectiveness of the proposed method.
\end{itemize}

\section{Related Work}
\label{sec:related}

\begin{figure*}
\centering
\includegraphics[width=0.92\linewidth]{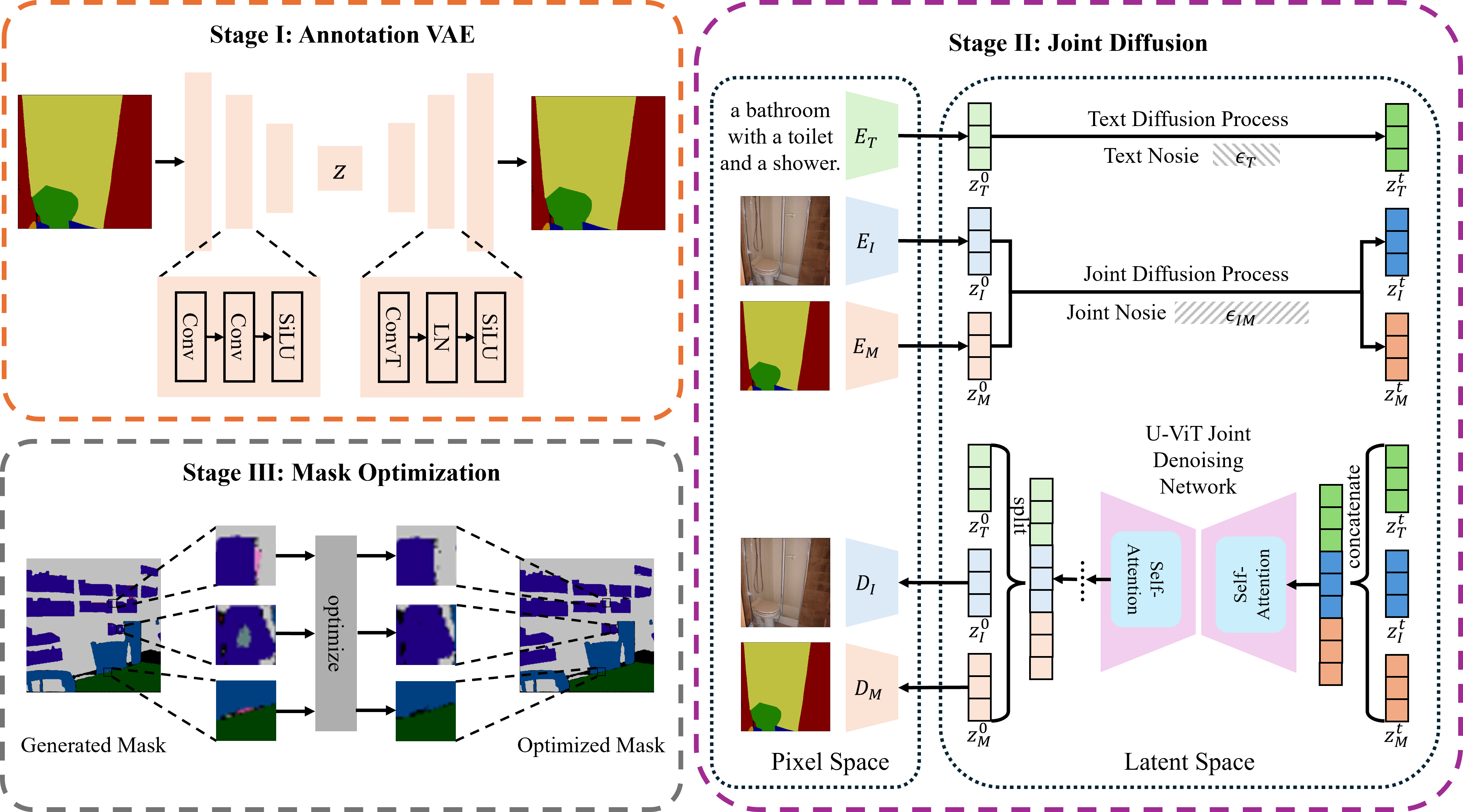}
\caption{Three stages of JoDiffusion. We first train an annotation VAE to efficiently encode sparse and discrete category maps into a compact latent space for diffusion model alignment. Next, we jointly model the relationship between text, images, and pixel-level annotation masks to enhance the semantic consistency of the generated results. Finally, we optimize the generated annotation masks to improve the quality of semantic segmentation results.}
\label{fig:method}
\end{figure*}

\subsection{Text-to-image Diffusion Models}

Diffusion models have made breakthrough progress in the field of image generation in recent years. 
% Compared with generative adversarial networks~\cite{gan}, diffusion models have the advantages of more stable training and higher generation quality. The core idea of the diffusion model is to generate high-quality samples by adding gradually increasing Gaussian noise to the data, mapping the data to the noise distribution, and training a denoising network to gradually restore the original data.
Early diffusion models~\cite{ddpm,ddim} achieved high-fidelity image generation through simple forward denoising and reverse denoising processes. Subsequently, models such as DALL·E~\cite{dalle,dalle2,dalle3} and Imagen~\cite{imagen,imagen3} adopted cross-modal conditional generation methods to apply diffusion models to text-to-image generation tasks, and surpassed GANs~\cite{gan} in terms of image clarity and semantic consistency. In order to improve inference efficiency, latent diffusion models~\cite{sd,sdxl,sd3} perform diffusion modeling in the latent space, greatly reducing the computational complexity while maintaining high-quality generation capabilities, which has promoted the popularity of diffusion models.
Subsequently, conditional image generation methods~\cite{controlnet,ipadapter,unicontrolnet} introduced additional control in the diffusion model to make the generation process more controllable. Multimodal generation methods~\cite{versatile,unidiffuser} jointly model the joint distribution of different modalities, allowing information such as text and images to interact with each other during the diffusion process, thereby achieving bidirectional control generation.

% Despite these advancements, existing text-to-image models primarily focus on generating images directly from text, with limited attention to pixel-level annotation generation for semantic segmentation. In contrast, JoDiffusion pioneers a novel joint-generation paradigm, where images and pixel-wise annotation masks are synthesized simultaneously within a unified framework, ensuring semantic consistency and scalability.

\subsection{Semantic Dataset Generation}
Early attempts at semantic segmentation dataset generation leveraged GAN-based models~\cite{datasetgan,bigdatasetgan}, where semantic information was extracted from the latent space, and pixel-level annotations were inferred using additional decoders. With the emergence of diffusion models~\cite{ddpm,ddim} demonstrating superior image synthesis quality, recent efforts have shifted toward diffusion-based dataset generation.
Existing approaches can be categorized into two main pipelines: Image2Mask and Mask2Image. The Image2Mask pipeline first generates images using a diffusion model, and infers the corresponding pixel-level annotation masks by parsing the features or attention maps in the generation process. For example, DiffuMask~\cite{diffumask} extracts category-related salient areas from the image generation process by analyzing the cross-attention mechanism of the diffusion model, and further infers annotation masks using Affinity Net. Dataset Diffusion~\cite{datasetdiffusion} is optimized on this basis, combining a large language model~\cite{gpt4} to generate more diverse text descriptions, and using self-attention maps to improve the quality of semantic masks. SDS~\cite{sds} further introduces perturbation-based CLIP similarity and class-balance annotation similarity to filter the generated images to reduce data noise and improve the effectiveness of the dataset.
In contrast, the Mask2Image method generates the corresponding images through the diffusion model based on the semantic masks. For example, FreeMask~\cite{freemask} uses the mask-to-image generation method FreestyleNet~\cite{freestylenet}, and designs a series of filtering strategies to suppress erroneously synthesized areas to ensure the quality of generated data. SegGen~\cite{seggen} train an additional text-to-mask model to make the generated semantic masks more diverse, thereby improving the generalization ability of the semantic segmentation model.

% While both paradigms contribute to dataset generation, they rely on sequential generation steps, leading to challenges in maintaining semantic consistency and scalability. JoDiffusion overcomes these limitations by jointly modeling images and annotation masks, enabling end-to-end synthesis of diverse and semantically coherent datasets from text prompts alone.

% \begin{figure*}[ht]
% \begin{minipage}{.55\linewidth}
% \centering
% \includegraphics[width=\linewidth]{3.png}
% \end{minipage}
% \hfill
% \begin{minipage}{.35\linewidth}
% \centering
% \begin{tabular}{lc}
% \toprule
% \makebox[0.4\textwidth][l]{\textbf{Dataset}} & \makebox[0.3\textwidth][c]{\textbf{mIoU $ \uparrow $ }} \\
% \midrule
% Pascal VOC & 99.50 \\
% MS COCO & 98.85 \\
% ADE20K & 98.74 \\
% \bottomrule
% \end{tabular}
% \end{minipage}
% \caption{Left: visualization of reconstructed pixel-level annotation masks on the validation sets. The first line is the input and the second line is the reconstruction result. Right: reconstruction mIoU of pixel-level annotation masks on the validation sets.}
% \label{fig:vae}
% \end{figure*}

\section{Method}
\label{sec:method}

\subsection{Problem Setup}

Our goal is to learn a joint generative model $ \mathcal{G}_\theta(I,M|T) $ that synthesizes images and corresponding annotation masks from text captions $ T $, using a real-world semantic segmentation dataset $ \mathcal{D}_{real}=\{(I_i,M_i)\}_{i=1}^{N_{real}} $ as supervision. The generated synthetic dataset $ \mathcal{D}_{syn}=\{(I_i,M_i)\}_{i=1}^{N_{syn}} $ should align with $ \mathcal{D}_{real} $ in terms of category distribution, object structures, and visual characteristics while introducing greater diversity to enhance the generalization of semantic segmentation models. Here, $ \theta $ represents the parameters of the generative model, and $ I_i, M_i $ denote the RGB image and its corresponding annotation mask, respectively. Finally, we evaluate our approach by training semantic segmentation models on $ \mathcal{D}_S $ and $ \mathcal{D}_R \cup \mathcal{D}_S $.

\begin{figure}[t]
\centering
\includegraphics[width=\linewidth]{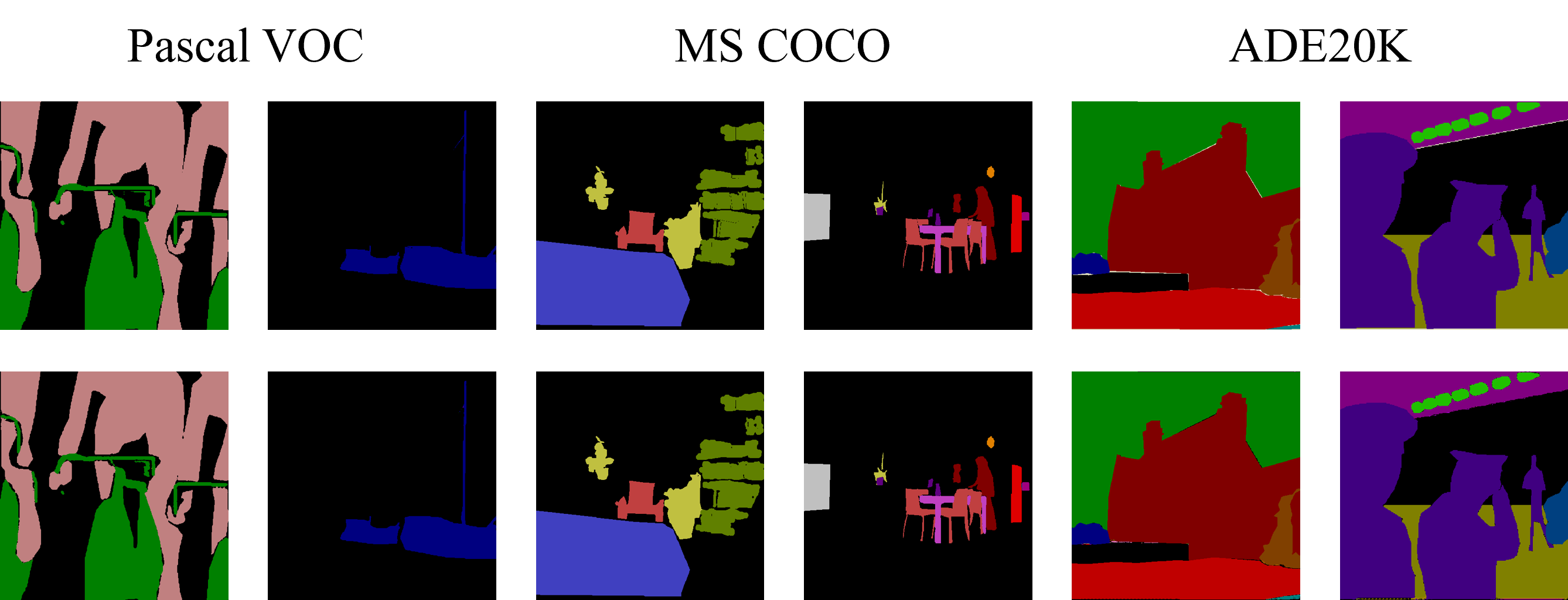}
\hfill
\caption{visualization of reconstructed pixel-level annotation masks on the validation sets. The first line is the input and the second line is the reconstruction result.}
\label{fig:vae}
\end{figure}

% Our objective is to use the real-world semantic segmentation dataset $ \mathcal{D}_{real}=\{(I_i,M_i),i=1,...,N_{real}\} $ to learn joint images and annotation masks generative model $ \mathcal{G}_\theta(I,M|T) $ based on text caption $ T $ to generate a synthetic dataset $ \mathcal{D}_{syn}=\{(I_i,M_i),i=1,...,N_{syn}\} $, which is consistent with the real dataset $ \mathcal{D}_{real} $ in terms of category distribution, object structure and visual characteristics, and can provide more diverse scenarios to enhance the generalization ability of the semantic segmentation model, where $ \theta $ is the parameter of the generative model, $ I_i $ and $ M_i $ are RGB image and its corresponding annotation masks. Finally, we use $ \mathcal{D}_S $ and $ \mathcal{D}_R \cup \mathcal{D}_S $ to train semantic segmentation models respectively to evaluate the effectiveness of the method.

\subsection{Overview}

As shown in Fig.~\ref{fig:method}, our method consists of three key stages: 1) Annotation VAE training: we first train an annotation VAE network to encode the annotation masks to obtain a compact latent representation. 
% This not only reduces redundant information, but also improves the stability of the diffusion model in the mask generation process. 
2) Joint diffusion modeling: we train the diffusion model based on text, images, and annotation masks to jointly model the relationship between the image latent variable $ z_{I} $ and the mask latent variable $ z_{M} $ in the latent space, and guide them through the text condition $ T $. In each denoising process, the model not only reconstructs the image features, but also ensures that the category of the annotation masks is consistent with the image content.
3) Mask optimization strategy: since the diffusion process may introduce label inconsistencies in small regions, we post-process them by using the majority class of its edge pixels, to correct the pixel-level annotations and optimize the final segmentation quality. The resulting dataset is used to train the semantic segmentation model.

\subsection{Annotation VAE}

To enable joint text-based generation of images and annotation masks, we adopt a latent diffusion model~\cite{uvit,unidiffuser}, where the image encoder maps RGB images into a latent space. To maintain consistency with this setup, we introduce an Annotation VAE to encode annotation masks into a corresponding latent representation.

% The semantic mask $ M $ and the RGB image $ I $ exhibit significant differences in data characteristics. Specifically, RGB images contain continuous pixel values, formally expressed as, $ I \in [0,255]^{H \times W \times 3} $, where $ H $ and $ W $ denote the image height and width, respectively. In contrast, The semantic mask represents a discrete category mapping, defined as $ M \in \{0,1,...,N_C\}^{H \times W}$, where $ N_C $ is the total number of categories, and each pixel is assigned a single category label. Directly applying an image VAE for encoding may result in invalid category values in the reconstructed pixel-level annotations. To address this, we propose an annotation VAE to learn a latent representation of the pixel-level annotations, enhancing its stability in the subsequent diffusion model.

Annotation masks are typically stored as single-channel category indices. Directly normalizing and feeding it into VAE may result in adjacent category values being too close, making it challenging for the model to accurately differentiate them. In order to improve the category discrimination and reduce the computational overhead, we employ binary encoding as the input representation of the annotation VAE. Specifically, the category of each pixel $ M(i,j) $ is converted into a binary representation $  M_{\text{bin}} $.
% of $ d=\lceil\log_2 N_C\rceil $ channels.
% , formulated as:
% \begin{equation}
%     M_{\text{bin},k(i,j)} = \lfloor \frac{M(i,j)}{2^k}\rfloor\mod 2, k = 0,1,...,d-1.
% \end{equation}

% This transformation yields the encoded mask representation $ M_{\text{bin}} \in \{0,1\}^{H \times W \times d} $, which enhances category separation and prevents adjacent categories from having overly similar values after normalization. Additionally, it significantly reduces the number of channels compared to one-hot encoding, thereby improving VAE training efficiency.

Annotation VAE follows a lightweight architecture comprising of an encoder $ E_M $ and a decoder $ D_M $, both utilizing a small number of convolutional and transposed convolutional layers. Compared to the image VAE used in the diffusion model, the annotation VAE not only significantly reduces the number of parameters ($ \approx $ 50M $ vs. $ 300M), as shown in Fig.~\ref{tab:vae}, but also maintains high reconstruction quality.

Since annotation VAE serves purely as a compression tool rather than a generative model, we do not impose a standard normal prior on its latent variables. Consequently, KL divergence regularization is omitted, and the model is trained solely using cross-entropy loss, defined as:
\begin{equation}
    \mathcal{L}_{\text{Annotation VAE}} = -\sum_{(i,j)} \sum_{c=0}^{N_C} M_{\text{one-hot},(i,j,c)} \log \bar{M}_{(i,j,c)},
\end{equation}
where $ M_{\text{one-hot},(i,j,c)} $ represents the ground truth one-hot category at pixel $ 
(i,j) $, and $ \bar{M}_{(i,j,c)} $ is the predicted probability obtained from the softmax output of the decoder. After training, given the latent representation $ z_M $ encoded by $ E_M $, the reconstructed semantic mask is obtained by applying an argmax operation over the softmax output of the decoder: $ \hat{M} = \arg\max (D_M(z_M)) $.

\begin{figure*}
\centering
\includegraphics[width=0.9\linewidth]{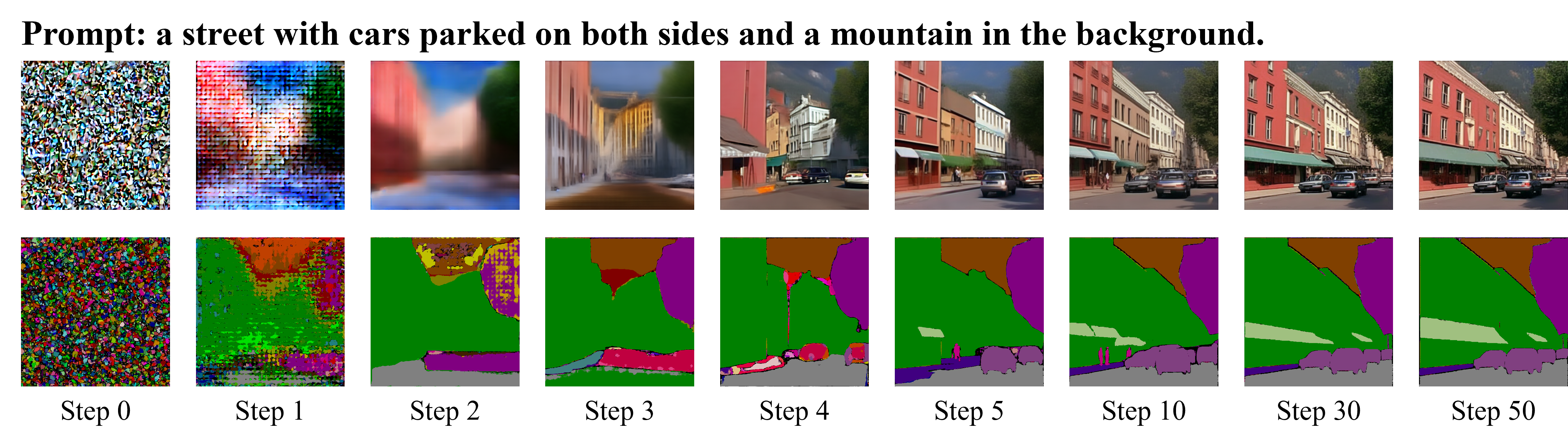}
\caption{Visualization of joint generation result at different timesteps. A color map is applied for better visualization.}
\label{fig:denoise}
\end{figure*}

\begin{table}[t]
\centering
\begin{tabular}{l|c}
\toprule
\makebox[0.15\textwidth][l]{\textbf{Dataset}} & \makebox[0.15\textwidth][c]{\textbf{mIoU $ \uparrow $ }} \\
\midrule
Pascal VOC & 99.50 \\
MS COCO & 98.85 \\
ADE20K & 98.74 \\
\bottomrule
\end{tabular}
\caption{Reconstruction mIoU of pixel-level annotation masks on the validation sets.}
\label{tab:vae}
\end{table}

\subsection{Joint Diffusion}

To ensure that the generated image and its corresponding pixel-level annotation masks remain semantically consistent, we adopt a joint diffusion process that models their shared distribution. Unlike Image2Mask pipeline, which first generates an image and infers its annotation masks, or Mask2Image pipeline, which generates annotation masks and then conditions the image generation, our approach diffuses and denoises images and annotation masks simultaneously. This bidirectional feature interaction allows for richer semantic alignment and improved scalability.

Our method builds upon Unidiffuser~\cite{unidiffuser}.
%, based on the denoising diffusion probability model~\cite{ddpm} framework. 
Compared to methods like SDXL~\cite{sdxl}, which rely on cross-attention to model text-image relationships, it concatenates text and image features and applies self-attention to model them, which offers greater flexibility for tuning. Specifically, given an image $ I $, we first generate a descriptive caption $ T $ using BLIP-2~\cite{blip2}. We then use the CLIP~\cite{clip} text encoder $ \mathcal{E}_T $, image encoder $ \mathcal{E}_I $, and the image VAE $ E_I $ encode them into latent space:
\begin{equation}
    z_T = \mathcal{E}_T(T),\quad z_I = [\mathcal{E}_I(I), E_I(I)].
\end{equation}

To integrate annotation masks $ M $ into this process, we leverage the Annotation VAE trained in the previous stage to obtain their latent representation: $ z_M = E_M(M) $. To ensure consistency between images with annotation masks, we diffuse $ z_I $ and $ z_M $ jointly instead of treating them as independent diffusion processes. We achieve this by introducing a shared noise perturbation $ \epsilon_{IM} $, maintaining semantic alignment during diffusion.

The forward process progressively injects Gaussian noise into $ z_I^0 $ and $ z_M^0 $, simulating a degradation path that enables effective denoising:
\begin{equation}
q(z_I^t, z_M^t|z_I^0, z_M^0) = \mathcal{N}(
\sqrt{\bar{\alpha}_t}\begin{bmatrix} z_I^0 \\ z_M^0 \end{bmatrix}, (1 - \bar{\alpha}_t)I),
\end{equation}
where $ z_I^0 = z_I, z_M^0 = z_M $ and $ \bar{\alpha}_t $ controls the noise schedule at timestep $ t $. This formulation ensures that both the image and annotation masks share the same noise perturbation $ \epsilon_{IM} \sim \mathcal{N}(0,I) $, maintaining structural consistency during training.

To recover the original image and pixel-level annotation masks pair from the noisy latent variables $ (z_I^t, z_M^t) $, we model the joint denoising distribution: 
\begin{equation}
p_\theta(z_I^{t-1}, z_M^{t-1}|z_I^t, z_M^t, z_T) = \mathcal{N}(\mu_\theta(z_I^t, z_M^t, z_T, t), \sigma^2_tI),
\end{equation}
where $ \sigma^2_t $ is determined by the predefined noise schedule and controls the level of randomness at each denoising step. The denoised mean $ \mu_{\theta} $ captures the underlying relationship between the image and pixel-level annotation masks:
\begin{small}
\begin{equation}
\mu_\theta(z_I^t, z_M^t, z_T, t) = \frac{1}{\sqrt{\alpha_t}}
\left(
\begin{bmatrix} z_I^t \\ z_M^t \end{bmatrix} - \frac{1 - \alpha_t}{\sqrt{1 - \bar{\alpha}_t}}\epsilon_\theta(z_I^t, z_M^t, z_T, t)
\right)
\end{equation}
\end{small}
where $ \epsilon_\theta(z_I^t, z_M^t, z_T, t) $ is the denoising network, which predicts the noise added during the forward diffusion process. Instead of estimating independent noise components, the network learns a joint representation, leveraging shared information between the image and annotation masks.

The model is trained using the standard mean squared error loss, where the image and annotation masks part is:
\begin{equation}
\mathcal{L}_{\text{denoising}} = \mathbb{E}_{t, z_I^0, z_M^0, \epsilon}
\left[ \| \epsilon_\theta(z_I^t, z_M^t, z_T, t) - \epsilon_{IM} \|^2 \right],
\end{equation}
where $ \epsilon_{IM} $ is the noise that was added during the forward diffusion process. By minimizing it, the model effectively denoises latent representations while preserving semantic integrity between the image and annotation masks. This reinforces semantic alignment in generated pairs, leading to improved performance in downstream segmentation tasks.

\begin{figure*}
\centering
\includegraphics[width=0.85\linewidth]{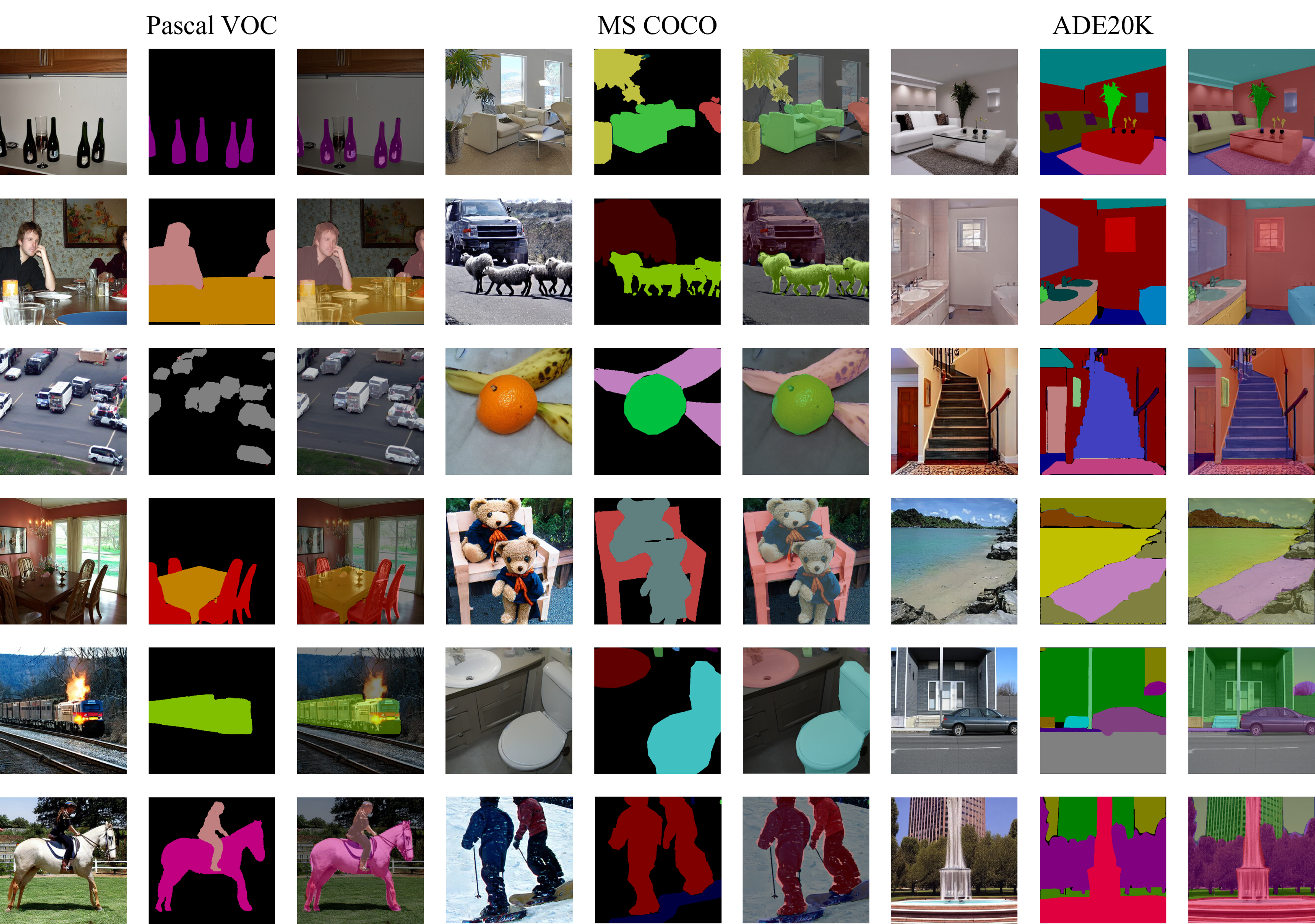}
\caption{Visualization of joint generation result of three datasets. A color map is applied for better visualization.}
\label{fig:vis}
\end{figure*}

\subsection{Mask Optimization}

While the joint diffusion process ensures semantic consistency between images and annotation masks, the pixel-level annotations generated by the model may still contain noise, especially around small target areas and object boundaries. This noise can manifest as speckle or label inconsistencies, which often lead to local deviations in labels and degrade the performance of downstream segmentation tasks. To address this, we propose a boundary mode-based mask optimization strategy. This method analyzes label distribution of boundary pixels and corrects small regions by replacing their labels with the most frequent category in that region, thus enhancing label consistency and suppressing noise.

% The pixel-level annotations generated by the diffusion model may contain speckle noise or label inconsistencies due to imperfections in the denoising process, especially in small target areas and object boundaries. These artifacts can lead to local label deviations, thereby degrading segmentation performance. To mitigate this issue, we propose a boundary mode-based mask optimization strategy, which analyzes the label distribution of boundary pixels and corrects small regions by replacing their labels with the most frequent boundary category (i.e., mode). This enhances label consistency while suppressing noise.

Let $ R \subset \{1,...,H\} \times \{1,...,W\} $ denote a small target region in the annotation mask, satisfying $ |R| < \tau $, where $ |R| $ being the number of pixels in $ R $, and $ \tau $ is a dataset-dependent threshold, typically set to identify small objects or noise regions. Small regions are particularly prone to noise, necessitating targeted refinement. To correct the labels in $ R $, we first define its boundary pixel set as $ \hat{R} $ and compute the mode of the label values among these boundary pixels:
\begin{equation}
c^\ast = \arg\max_c\sum_{(i,j)\in \hat{R}} \mathbb{I} (x_{i,j}=c),
\end{equation}
where $ x_{i,j} $ is the label at pixel $ (i,j) $, and $ \mathbb{I}(\cdot) $ is an indicator function that counts occurrences of category in the boundary pixels. The calculated mode $ c^\ast $ represents the most frequent category in $ \hat{R} $, which is then used to reassign all pixels in $ R $:
\begin{equation}
\forall (i,j) \in R, \quad x_{i,j} \leftarrow c^\ast. 
\end{equation}

The effectiveness of this correction method is grounded in statistical estimation principles.  Given a small target region $ R $, its true category label may be ambiguous due to noise introduced in the diffusion process. However, the boundary pixels $ \hat{R} $ are more likely to retain correct labels due to the inherent continuity of semantic regions in natural images. This assumption is supported by two key observations: adjacent pixels in real-world images typically belong to the same category, and errors introduced by the diffusion process tend to be randomly distributed in small isolated regions rather than along structured object boundaries.

Under these assumptions, the mode $ c^\ast $ of the boundary labels provides a reliable estimate of the true category of the target region. From a statistical perspective, this process can be viewed as a maximum likelihood estimation, where the most frequent category among the boundary pixels serves as the most probable label assignment for $ R $. Formally, this can be expressed as:
\begin{equation}
c^\ast = \arg\max_c P(c | \hat{R}),
\end{equation}
where $ P(c | \hat{R}) $ represents the empirical distribution of labels in the boundary region. Assuming an approximately uniform prior over categories, this estimation reduces to selecting the mode of the boundary labels. By replacing the labels in $ R $ with $ c^\ast $, we effectively minimize the probability of incorrect category assignments while preserving structural coherence in the segmentation mask.

\begin{table*}[t]
\small
\centering
\begin{tabular}{ccccccccc}
\toprule
\multirow{3.5}{*}{\textbf{Segmenter}} & \multirow{3.5}{*}{\textbf{Backbone}} & \multirow{3.5}{*}{{\textbf{Method}}} & \multicolumn{3}{c}{\textbf{Pascal VOC}} & \multicolumn{3}{c}{\textbf{MS-COCO}} \\
\cmidrule(lr){4-6} \cmidrule(lr){7-9}
&  &  & \makebox[0.06\textwidth][c]{\textbf{\makecell[c]{Data \\ Size}}} & \makebox[0.08\textwidth][c]{\textbf{\makecell[c]{mIoU \\ (Syn)}}} & \makebox[0.08\textwidth][c]{\textbf{\makecell[c]{mIoU \\(Real+Syn)}}} & \makebox[0.06\textwidth][c]{\textbf{\makecell[c]{Data \\ Size}}} & \makebox[0.08\textwidth][c]{\textbf{\makecell[c]{mIoU \\ (Syn)}}} & \makebox[0.08\textwidth][c]{\textbf{\makecell[c]{mIoU \\ (Real+Syn)}}} \\
\midrule
\multirow{9}{*}{DeepLabV3} & \multirow{4.5}{*}{ResNet50} & Raw Dataset & $11.5k$ & \multicolumn{2}{c}{77.4} & $118k$ & \multicolumn{2}{c}{48.9} \\
\cmidrule{3-9}
&  & SDS & $26k$ & 60.4 & 77.6 & $50k$ & 31.0 & 50.3 \\
&  & Dataset Diffusion & $40k$ & 61.6 & 77.6 & $80k$ & 32.4 & 54.6 \\
&  & JoDiffusion & $40k$ & \textbf{72.5} & \textbf{78.3} & $80k$ & \textbf{42.6} & \textbf{56.4} \\
\cmidrule{2-9}
& \multirow{4.5}{*}{ResNet101} & Raw Dataset & $11.5k$ & \multicolumn{2}{c}{79.9} & $118k$ & \multicolumn{2}{c}{54.9} \\
\cmidrule{3-9}
&  & SDS & $26k$ & 59.1 & 79.8 & $50k$ & 31.8 & 56.8 \\
&  & Dataset Diffusion & $40k$ & 64.8 & 80.3 & $80k$ & 34.2 & 57.4 \\
&  & JoDiffusion & $40k$ & \textbf{75.8} & \textbf{80.7} & $80k$ & \textbf{44.9} & \textbf{59.1} \\
\midrule
\multirow{5.5}{*}{Mask2Former} & \multirow{5.5}{*}{ResNet50} & Raw Dataset & $11.5k$ & \multicolumn{2}{c}{77.3} & $118k$ & \multicolumn{2}{c}{57.8} \\
\cmidrule{3-9}
&  & DiffuMask & $60k$ & 57.4 & 77.5 & - & - & - \\
&  & SDS & $26k$ & 59.8 & 78.1 & 50k & 29.8 & 57.7 \\
&  & Dataset Diffusion & $40k$ & 60.2 & 78.2 & $80k$ & 31.0 & 57.8 \\
&  & JoDiffusion & $40k$ & \textbf{74.5} & \textbf{79.4} & $80k$ & \textbf{44.6} & \textbf{58.5} \\
\bottomrule
\end{tabular}
\caption{Comparisons in mIoU with Image2Mask methods on Pascal VOC and MS-COCO dataset.}
\label{tab:img2mask-result}
\end{table*}

\begin{table}[ht]
\small
\centering
\begin{tabular}{cccccc}
\toprule
\multirow{3.5}{*}{\textbf{Backbone}} & \multirow{3.5}{*}{{\textbf{Method}}} & \multicolumn{2}{c}{\textbf{Pascal VOC}} & \multicolumn{2}{c}{\textbf{ADE20K}}\\
\cmidrule(lr){3-4} \cmidrule(lr){5-6}
&  & \makebox[0.03\textwidth][c]{\textbf{\makecell[c]{Data \\ Size}}} & \makebox[0.04\textwidth][c]{\textbf{\makecell[c]{mIoU}}} & \makebox[0.03\textwidth][c]{\textbf{\makecell[c]{Data \\ Size}}} & \makebox[0.04\textwidth][c]{\textbf{\makecell[c]{mIoU}}} \\
\midrule
\multirow{4.5}{*}{ResNet50} & Raw Data & $11.5k$ & 77.3 & $20k$ & 47.2 \\
\cmidrule{2-6}
& SegGen & - & - & $ 1 $M  & \textbf{49.9} \\
& FreeMask & 40k & \underline{77.9}$ ^\dagger $ & $ 40k $ & 48.2$ ^\dagger $ \\
& JoDiffusion & 40k & \textbf{79.4} & $ 40k $ & \underline{48.4} \\
\midrule
\multirow{3.5}{*}{Swin-S} & Raw Data & $11.5k$ & 83.8 & $20k$ & 51.6 \\
\cmidrule{2-6}
& FreeMask & 40k & \underline{84.2}$ ^\dagger $ & $ 40k $ & \underline{52.1}$ ^\dagger $ \\
& JoDiffusion & 40k & \textbf{85.1} & $ 40k $ & \textbf{52.2} \\
\bottomrule
\end{tabular}
\caption{Comparisons in mIoU with Mask2Image methods on ADE20K dataset. $ \dagger $ means our reproduced results.}
\label{tab:mask2-result}
\end{table}

\section{Experiments}
\label{sec:exp}

\subsection{Datasets}

\textbf{Pascal VOC}~\cite{voc} is a widely used benchmark for semantic segmentation, containing 20 object categories and 1 background category. Following previous work~\cite{datasetdiffusion,sds}, we incorporate the Semantic Boundaries Dataset~\cite{sbd} to extend its training set, resulting in 10,582 training images and 1,449 validation images. \textbf{MS-COCO}~\cite{coco} provides a more diverse and complex dataset with 80 object categories and one background category. It consists of 118,288 training images and 5,000 validation images, offering a challenging setting with high intra-class variance and occlusions. \textbf{ADE20K}~\cite{ade20k} is a scene parsing dataset containing 150 fine-grained semantic concepts, making it suitable for dense prediction tasks. It includes 20,210 training images and 2,000 validation images, covering a broad range of indoor and outdoor scenes.

\subsection{Implementation Details}

For all datasets, we resize images and annotation masks to 512 $ \times $ 512 512 for training both VAE and diffusion models. We use the AdamW~\cite{adamw} optimizer for all training stages, and apply random horizontal flipping as data augmentation. Additional architecture details and hyperparameters are provided in the supplementary material.

% \textbf{VAE Training Settings.} The annotation VAE is trained with a weight decay of 0.05 and a max gradient norm of 3. The learning rates are set to 1e-5, 1e-4, and 1e-4 for Pascal VOC, MS-COCO, and ADE20K datasets, respectively.

% \textbf{Diffusion Model Training Settings.} We train the diffusion model with a fixed learning rate of 5e-5, weight decay of 0.01, and a max gradient norm of 1.

% \textbf{Joint Generation Setting.} We use captions from the training images of each dataset as text prompts. Then, we employ the DPMSolverMultistep scheduler to sample from the trained diffusion model, using 50 denoising steps. 
% % And finally apply our mask optimization strategy with a threshold of $ \tau = 20 $.

% \textbf{Semantic Segmentor Training Settings.} We use the MM-Segmentation framework~\cite{mmseg} for training. On Pascal VOC and MS-COCO, we follow Image2Mask settings and train the segmenter exclusively on synthetic data. On ADE20K, we adopt the Mask2Image strategy and jointly train the segmenter using both real and synthetic data.

% \textbf{Evaluation Metric.} We use mean intersection over union (mIoU) to evaluate both VAE reconstruction quality and semantic segmentation performance. All reported mIoU scores are obtained using single-scale testing.

% Additional architecture details and hyperparameters are provided in the supplementary material.

\subsection{Comparison with State-of-the-Arts}

We compare our method with the state-of-the-art Image2Mask methods, including DiffuMask~\cite{diffumask}, Dataset Diffusion~\cite{datasetdiffusion} and SDS~\cite{sds}, as well as Mask2Image methods such as FreeMask~\cite{freemask} and SegGen~\cite{seggen}. To ensure a fair comparison, we reproduced FreeMask using the same amount of data as ours and applied the same filtering strategy to our method.

\subsubsection{Qualitative Results}

Fig.~\ref{fig:vae} illustrates the reconstruction performance of our annotation VAE. The first row presents the ground-truth pixel-level annotation masks, while the second row shows the reconstructed annotations after encoding and decoding. Our model effectively preserves the structural and categorical details of the original annotations, achieving high reconstruction fidelity with minimal information loss.

Fig.~\ref{fig:denoise} visualizes the intermediate diffusion steps of our joint generation process. Starting from an initial noisy representation, our method progressively refines both the image and its corresponding annotation masks, improving structural and semantic coherence over time. As diffusion progresses, contours become sharper, and the generated annotations better align with object semantics, demonstrating the effectiveness of our joint generation strategy.

Fig.~\ref{fig:vis} presents the final synthesized image and annotations pairs on three datasets. For each dataset, the third column overlays the generated image and annotations for better visualization. The results indicate that our approach not only produces high-quality images across diverse datasets but also maintains strong semantic alignment between generated annotations and image content. Additional qualitative results can be found in the supplementary material.

\subsubsection{Quantitative Results}

Tab.~\ref{tab:vae} reports the mIoU of our trained annotation VAE on three datasets. Our method achieves reconstruction accuracy exceeding 98\%, demonstrating its effectiveness in compactly encoding annotation masks while preserving critical structural information.

Tab.~\ref{tab:img2mask-result} compares our method with Image2Mask approaches on Pascal VOC and MS-COCO datasets. Across multiple segmentation architectures and backbones, our approach significantly outperforms prior methods. % Particularly on Pascal VOC, the performance gap between our synthetic data and real data is notably reduced.

Tab.~\ref{tab:mask2-result} presents the results on Pascal VOC and ADE20K datasets, where we follow the Mask2Image paradigm by training the Mask2Former segmenter with both real and synthetic data. Our approach consistently outperforms existing Mask2Image methods across multiple backbones. Additional results can be found in the supplementary material.

\subsection{Discussion}

\textbf{Effectiveness of the mask optimization strategy.} 
% To validate the contribution of our mask optimization strategy, 
We analyze the effect of different regional thresholds $ \tau $ on segmentation performance on Psacal VOC dataset. As shown in Tab.~\ref{tab:ablation-mask}, applying mask optimization improves performance compared to the baseline without optimization.% $ \tau = 0 $.

\begin{table}[ht]
\small
\centering
\begin{tabular}{lcccc}
\toprule
\textbf{$ \tau $} & \textbf{$ \tau = 0 $} & \textbf{$ \tau = 20 $} & \textbf{$ \tau = 50 $} & \textbf{$ \tau = 100 $} \\
\midrule
mIoU $ \uparrow $ & 71.37 & \textbf{72.47} & 72.38 & 72.38 \\
\bottomrule
\end{tabular}
\caption{Results on different mask optimization threshold $ \tau $.}
\label{tab:ablation-mask}
\end{table}

\textbf{Effectiveness of the generated data size.} We investigate the impact of different amounts of generated training data on segmentation performance on Psacal VOC dataset. As shown in Tab.~\ref{tab:ablation-size}, increasing the dataset size consistently improves performance.

\begin{table}[!h]
\small
\centering
\begin{tabular}{lcccc}
\toprule
\textbf{Data Size} & \textbf{ $ 5k $ } & \textbf{ $ 10k $ } & \textbf{ $ 20k $ } & \textbf{ $ 40k $ } \\
\midrule
mIoU $ \uparrow $ & 68.54 & 70.02 & 70.97 & \textbf{72.47} \\
\bottomrule
\end{tabular}
\caption{Results on different data sizes.}
\label{tab:ablation-size}
\end{table}

\section{Conclusion}
\label{sec:conclusion}
In this paper, we introduce JoDiffusion, a novel framework for joint image and annotation mask generation framework. Unlike traditional Image2Mask and Mask2Image approaches, our method directly models the joint distribution of images and their corresponding annotation masks. By incorporating an annotation VAE and an effective mask optimization strategy, our approach significantly outperforms prior methods in segmentation performance on Pascal VOC, MS-COCO, and ADE20K, demonstrating its efficacy in generating high-quality synthetic segmentation data.

% Although our method is implemented with Unidiffuser~\cite{unidiffuser} for semantic segmentation, it remains architecture-agnostic and adaptable to various tasks. Future work will explore its extension to more advanced generative models and broader dense prediction tasks.

\section*{Acknowledgments}
This work is supported in part by the National Natural Science Foundation of China under Grand 62372379, Grant 62472359, and Grant 62472350; in part by the Xi’an’s Key Industrial Chain Core Technology Breakthrough Project: AI Core Technology Breakthrough under Grand 23ZDCYJSGG0003-2023; in part by National Key Laboratory of Science and Technology on Space-Born Intelligent Information Processing fundation under Grant TJ-04-23-04; in part by Innovation Foundation for Doctor Dissertation of Northwestern Polytechnical University under Grant CX2025092.

\bibliography{aaai2026}

\end{document}

\maketitle

\section{Annotation VAE Architecture}

The annotation VAE is designed for efficient encoding and reconstruction of pixel-level annotation masks. It focuses on providing a compact and effective model that performs well on large-scale segmentation datasets while maintaining low computational overhead. By using a compact latent space and binary-encoded categories, the model achieves high-quality reconstruction with a low number of parameters. This design allows for scalability and high performance in data generation and segmentation tasks, making it a powerful tool for large-scale applications. Below, we describe the key components of the architecture, including the encoder and decoder.

\subsection{Encoder}

The encoder transforms the input pixel-level annotation masks into a compressed latent representation. The process begins with an initial convolutional layer that applies a $3 \times 3$ kernel with padding to the input annotation masks, increasing its depth to match the first output block, which by default consists of 128 output channels. A SiLU~\cite{silu} activation is used to introduce non-linearity after this initial convolution.

Next, the encoder has three downsampling blocks that progressively reduce the spatial resolution while increasing the feature map depth. Each downsampling block includes a $3 \times 3$ convolution with padding to preserve spatial dimensions, followed by another $3 \times 3$ convolution with a stride of 2, which reduces the spatial resolution by half. SiLU activations are applied after each convolutional layer. The number of channels increases through the blocks, starting from 128 and progressing to 256, 512, and finally 512 channels, which allows the model to capture increasingly abstract features.

The encoder concludes with a GroupNorm~\cite{groupnorm} layer (32 groups) and a final convolution that maps the feature map to the latent space. The latent space is represented by a set of latent variables, with a default of 4 latent channels per annotation masks, and 2 latent variables per annotation masks. The encoder outputs a latent distribution, which is used for annotation masks reconstruction in the decoder.

\subsection{Decoder}

The decoder reconstructs the original annotation masks from the latent space representation. It begins with a Conv2d layer that projects the latent variables to a higher-dimensional feature space. This layer uses a $3 \times 3$ kernel and padding, with the number of input channels set to the latent channels and output channels set to the intermediate number of channels.

The decoder contains several upscaling blocks, which apply transposed convolutions to progressively upsample the feature maps to the original annotation masks resolution. Each upscaling block consists of a ConvTranspose2d layer to double the spatial resolution, followed by LayerNorm and SiLU activations for normalization and non-linearity. The number of upscaling blocks is set to 3 by default. These layers efficiently reconstruct the annotation masks by increasing the resolution of the feature maps.

The decoder concludes with a GroupNorm layer and a final $3 \times 3$ convolutional layer, producing the reconstructed annotation masks.

\section{Detailed Hyperparameters}

We provide detailed hyperparameters for training the annotation VAE, diffusion model, and downstream semantic segmentation tasks, as well as for the image and pixel-level annotation masks joint generation process on the Pascal VOC~\cite{voc}, MS-COCO~\cite{coco}, and ADE20K~\cite{ade20k} datasets. The corresponding settings are summarized in Tab.~\ref{tab:vae train}-\ref{tab:segmentation train}, respectively.

For training the annotation VAE, we use dataset-specific learning rates to ensure stable optimization across different segmentation datasets. The diffusion model is trained with a fixed learning rate and gradient clipping to enhance training stability. For downstream segmentation training, we mostly adhere to MMSegmentation~\cite{mmseg} defaults, listing only the modified hyperparameters.

\begin{table*}[htbp]
\centering
\begin{tabular}{lccc}
\toprule
\makebox[0.2\textwidth][l]{\textbf{Hyperparameter}} & \makebox[0.2\textwidth][c]{\textbf{Pascal VOC}} & \makebox[0.2\textwidth][c]{\textbf{MS-COCO}} & \makebox[0.2\textwidth][c]{\textbf{ADE20K}} \\
\midrule
in channels & 5 & 7 & 8 \\
intermediate channels & 512 & 512 & 512 \\
out channels & 21 & 81 & 151 \\
block out channels & (128, 256, 512, 512) & (128, 256, 512, 512) & (128, 256, 512, 512) \\
latent channels & 4 & 4 & 4 \\
resolution & 512 & 512 & 512 \\
training augmentation & \makecell[c]{Resize, \\ RandomCrop, \\ RandomHorizontalFlip} & \makecell[c]{Resize, \\ RandomCrop, \\ RandomHorizontalFlip} & \makecell[c]{Resize, \\ RandomCrop, \\ RandomHorizontalFlip} \\
validation augmentation & \makecell[c]{Resize, \\ CenterCrop} & \makecell[c]{Resize, \\ CenterCrop} & \makecell[c]{Resize, \\ CenterCrop} \\
batch size & 32 & 32 & 32 \\
epoch & 50 & 20 & 100 \\
optimizer & AdamW & AdamW & AdamW \\
learning rate & 1e-5 & 1e-4 & 1e-4 \\
lr scheduler & constant & constant & constant \\
AdamW - $ \beta_1 $ & 0.9 & 0.9 & 0.9 \\
AdamW - $ \beta_2 $ & 0.999 & 0.999 & 0.999 \\
AdamW - $ \epsilon $ & 1e-8 & 1e-8 & 1e-8 \\
AdamW - weight decay & 5e-2 & 5e-2 & 5e-2 \\
max grad norm & 3 & 3 & 3 \\
mixed precision & fp16 & fp16 & fp16 \\
\bottomrule
\end{tabular}
\caption{Training hyperparameters of annotation VAE.}
\label{tab:vae train}
\vspace{1cm}
\end{table*}

% \vspace{4cm}

\begin{table*}[htbp]
\centering
\begin{tabular}{lccc}
\toprule
\makebox[0.2\textwidth][l]{\textbf{Hyperparameter}} & \makebox[0.2\textwidth][c]{\textbf{Pascal VOC}} & \makebox[0.2\textwidth][c]{\textbf{MS-COCO}} & \makebox[0.2\textwidth][c]{\textbf{ADE20K}} \\
\midrule
caption & BLIP-2~\cite{blip2} & COCO Caption & BLIP-2~\cite{blip2} \\
resolution & 512 & 512 & 512 \\
training augmentation & \makecell[c]{Resize, \\ CenterCrop, \\ RandomHorizontalFlip} & \makecell[c]{Resize, \\ CenterCrop, \\ RandomHorizontalFlip} & \makecell[c]{Resize, \\ CenterCrop, \\ RandomHorizontalFlip} \\
sequence length & 2129 & 2129 & 2129 \\
noise type & joint & joint & joint \\
prediction type & epsilon & epsilon & epsilon \\
batch size & 32 & 32 & 32 \\
epoch & 100 & 20 & 100 \\
optimizer & 8bit AdamW & 8bit AdamW & 8bit AdamW \\
learning rate & 5e-5 & 5e-5 & 5e-5 \\
lr scheduler & constant & constant & constant \\
AdamW - $ \beta_1 $ & 0.9 & 0.9 & 0.9 \\
AdamW - $ \beta_2 $ & 0.999 & 0.999 & 0.999 \\
AdamW - $ \epsilon $ & 1e-8 & 1e-8 & 1e-8 \\
AdamW - weight decay & 1e-2 & 1e-2 & 1e-2 \\
max grad norm & 1 & 1 & 1 \\
mixed precision & fp16 & fp16 & fp16 \\
\bottomrule
\end{tabular}
\caption{Training hyperparameters of diffusion model.}
\label{tab:diffusion train}
\end{table*}

\begin{table*}[htbp]
\centering
\begin{tabular}{lccc}
\toprule
\makebox[0.2\textwidth][l]{\textbf{Hyperparameter}} & \makebox[0.2\textwidth][c]{\textbf{Pascal VOC}} & \makebox[0.2\textwidth][c]{\textbf{MS-COCO}} & \makebox[0.2\textwidth][c]{\textbf{ADE20K}} \\
\midrule
caption & BLIP-2~\cite{blip2} & COCO Caption & BLIP-2~\cite{blip2} \\
scheduler & DPMSolverMultistep & DPMSolverMultistep & DPMSolverMultistep \\
generate type & text2img & text2img & text2img \\
steps & 50 & 50 & 50 \\
Number of Samples Generated & 40,000 & 80,000 & 40,000 \\
precision & fp16 & fp16 & fp16 \\
optimization $ \tau $ & 20 & 20 & 20 \\
\bottomrule
\end{tabular}
\caption{Generation hyperparameters of diffusion model.}
\label{tab:diffusion generate}
\end{table*}

\begin{table*}[htbp]
\centering
\begin{tabular}{lccc}
\toprule
\makebox[0.2\textwidth][l]{\textbf{Hyperparameter}} & \makebox[0.2\textwidth][c]{\textbf{Pascal VOC}} & \makebox[0.2\textwidth][c]{\textbf{MS-COCO}} & \makebox[0.2\textwidth][c]{\textbf{ADE20K}} \\
\midrule
optimizer & AdamW & AdamW & AdamW \\
training type & \makecell[c]{synthetic only \\ real and synthetic} & \makecell[c]{synthetic only \\ real and synthetic} & real and synthetic \\
iterations & \makecell[c]{\textit{synthetic only:} \\ DepeplabV3-r50 20k \\ DepeplabV3-r101 20k \\ Mask2Former-r50 90k \\ \textit{real and synthetic:} \\ Mask2Former-r50 160k \\ Mask2Former-swin-t 160k \\ Mask2Former-swin-s 160k} & \makecell[c]{\textit{synthetic only:} \\ DepeplabV3-r50 80k \\ DepeplabV3-r101 80k \\ Mask2Former-r50 90k \\ \textit{real and synthetic:} \\ Mask2Former-r50 160k \\ Mask2Former-swin-t 160k \\ Mask2Former-swin-s 160k} & \makecell[c]{\textit{real and synthetic:} \\ Mask2Former-r50 320k \\ Mask2Former-swin-t 320k \\ Mask2Former-swin-s 320k} \\
\bottomrule
\end{tabular}
\caption{Training hyperparameters of semantic segmentation model.}
\label{tab:segmentation train}
\end{table*}

\section{Baselines}

To evaluate the effectiveness of our method, we compare it with state-of-the-art Image2Mask and Mask2Image approaches for semantic segmentation dataset generation. Below, we provide a detailed description of these baselines and outline our experimental settings for a fair comparison.

\subsection{Image2Mask Methods}

Image2Mask methods first generate images from text prompts and then infer their corresponding sematic masks through various heuristics or learned representations.

\begin{itemize}
    \item \textbf{DiffuMask}~\cite{diffumask} generates images using a diffusion model with a conditional class name, and uses a cross-attention map to obtain the sematic mask of the corresponding class according to the affinity net.
    \item \textbf{Dataset Diffusion}~\cite{datasetdiffusion} extends DiffuMask by incorporating multi-category text prompts and refining the generated sematic masks via self-training with an uncertainty-aware segmentation loss, improving mask quality.
    \item \textbf{SDS}~\cite{sds} further enhances Dataset Diffusion by introducing a perturbation-based CLIP similarity and a class-balance annotation similarity filter to filter out low-quality image-mask pairs, leading to a higher-fidelity synthetic dataset.
\end{itemize}

To ensure fair comparison, we adopt the experimental settings of these methods. Specifically, on the Pascal VOC~\cite{voc} and MS-COCO~\cite{coco} datasets, we use image captions derived from the training set as text prompts and generate the same number of image-mask pairs. These synthetic pairs are then exclusively used to train the semantic segmentation model, following the protocol of previous Image2Mask methods.

\subsection{Mask2Image Methods}

Mask2Image methods take sematic masks as input and generate realistic images that conform to the given structure while ensuring semantic consistency.

\begin{itemize}
    \item \textbf{FreeMask}~\cite{freemask} utilizes an image generator trained via FreestyleMask\cite{freestylenet} to synthesize images from training set masks. To improve mask accuracy, FreeMask applies a pre-trained segmentation model to identify and filter incorrect regions, followed by a hard-sample resampling strategy to enhance data diversity.
    \item \textbf{SegGen}~\cite{seggen} proposes a dual-model approach, consisting of a Text2Mask generator and a Mask2Image generator. The majority of its training data is produced by the Mask2Image model, with a 1:5 ratio between Text2Mask and Mask2Image data, making it heavily reliant on high-quality training set masks.
\end{itemize}

Following the settings of these methods, we train semantic segmentation models on the ADE20K~\cite{ade20k} dataset using both real and generated data. However, since SegGen is not open-source, we reproduce FreeMask under controlled conditions. Specifically, we reproduced FreeMask using the same amount of data as ours and applied the same filtering strategy to our method to ensure a fair comparison.

By maintaining identical data constraints and filtering procedures, we ensure that performance differences arise from the generation method itself rather than disparities in dataset size or pre-processing techniques.

\section{Additional Experimental Results}

To further validate the effectiveness of our proposed method, we provide additional experimental results in this section.
 
\subsection{Addition Comparison}

We conduct a more comprehensive comparison with the Mask2Image method under various backbone architectures. As shown in Tab.~\ref{tab:more-mask2-result}, FreeMask leverages additional semantic segmentation masks as generation conditions. While this leads to comparable performance with our method on the ADE20K dataset, which has more complex masks, our method significantly outperforms it on datasets with simpler mask structures such as VOC and COCO.
% In addition to the constrained setting where FreeMask is limited to 2,000 masks, we also compare against FreeMask when utilizing the entire training set’s masks for generation. As shown in Table 1, despite only using text as input, our method achieves results comparable to FreeMask, which relies on all available training set masks. This highlights the robustness of our approach in generating high-quality synthetic segmentation datasets without the need for pre-existing mask annotations.

\begin{table}[ht]
\centering
\begin{tabular}{llcc}
\toprule
\textbf{Backbone} & \textbf{Method} & \textbf{Data Size} & \textbf{mIoU $ \uparrow $} \\
\midrule
\multirow{11}{*}{ResNet50}& ADE20K     & $ 20k $           & 47.2  \\
                          & SegGen     & $ 20k $ + $ 1 $M  & \textbf{49.9}  \\
                          & FreeMask   & $ 20k $ + $ 40k $ & 48.2$ ^\dagger $ \\
                          & Ours       & $ 20k $ + $ 40k $ & \underline{48.4} \\
\cmidrule{2-4}
                          & VOC        & $ 11.5k $           & 77.3  \\
                          & FreeMask   & $ 11.5k $ + $ 40k $ & \underline{77.9}$ ^\dagger $ \\
                          & Ours       & $ 11.5k $ + $ 40k $ & \textbf{79.4} \\
\cmidrule{2-4}
                          & COCO       & $ 118k $           & 52.8  \\
                          & FreeMask   & $ 118k $ + $ 80k $ & \underline{54.0}$ ^\dagger $ \\
                          & Ours       & $ 118k $ + $ 80k $ & \textbf{58.5} \\
\midrule
\multirow{11}{*}{Swin-T}& ADE20K    & $ 20k $           & 48.7  \\
                          & FreeMask   & $ 20k $ + $ 400k$ & \textbf{52.0}  \\
                          & FreeMask   & $ 20k $ + $ 40k $ & \underline{50.4}$ ^\dagger $ \\
                          & Ours       & $ 20k $ + $ 40k $ & 50.3 \\
\cmidrule{2-4}
                          & VOC        & $ 11.5k $           & 81.3  \\
                          & FreeMask   & $ 11.5k $ + $ 40k $ & \underline{81.9}$ ^\dagger $ \\
                          & Ours       & $ 11.5k $ + $ 40k $ & \textbf{82.6} \\
\cmidrule{2-4}
                          & COCO       & $ 118k $           & 53.5  \\
                          & FreeMask   & $ 118k $ + $ 80k $ & \underline{57.4}$ ^\dagger $ \\
                          & Ours       & $ 118k $ + $ 80k $ & \textbf{59.5} \\
\midrule
\multirow{11}{*}{Swin-S}& ADE20K    & $ 20k $           & 51.6  \\
                          & FreeMask   & $ 20k $ + $ 400k$ & \textbf{53.3}  \\
                          & FreeMask   & $ 20k $ + $ 40k $ & 52.1$ ^\dagger $ \\
                          & Ours       & $ 20k $ + $ 40k $ & \underline{52.2} \\
\cmidrule{2-4}
                          & VOC        & $ 11.5k $           & 83.8  \\
                          & FreeMask   & $ 11.5k $ + $ 40k $ & \underline{84.2}$ ^\dagger $ \\
                          & Ours       & $ 11.5k $ + $ 40k $ & \textbf{85.1} \\
\cmidrule{2-4}
                          & COCO       & $ 118k $           & \underline{62.6}  \\
                          & FreeMask   & $ 118k $ + $ 80k $ & 61.3$ ^\dagger $ \\
                          & Ours       & $ 118k $ + $ 80k $ & \textbf{63.5} \\
\bottomrule
\end{tabular}
\caption{Comparisons in mIoU with Mask2Image methods on three dataset. $ \dagger $ means our reproduced results.}
\label{tab:more-mask2-result}
\end{table}

\subsection{Addition Discussions}

We further analyze the effect of different mask optimization thresholds $ \tau $ across datasets, as presented in Tab.~\ref{tab:ablation-mask-coco} and Tab.~\ref{tab:ablation-mask-ade20k}. As shown in the table, for datasets with more complex masks such as MS-COCO and ADE20K, mask optimization has a significant impact primarily when it is enabled or disabled, while the results remain robust to the specific choice of the threshold. Therefore, we set the threshold uniformly to $ \tau = 20 $ for all datasets.

\begin{table}[ht]
\centering
\begin{tabular}{lcccc}
\toprule
\textbf{$ \tau $} & \textbf{$ \tau = 0 $} & \textbf{$ \tau = 20 $} & \textbf{$ \tau = 50 $} & \textbf{$ \tau = 100 $} \\
\midrule
mIoU $ \uparrow $ & 42.32 & \textbf{42.57} & 42.55 & 42.38 \\
\bottomrule
\end{tabular}
\caption{Ablation results on mask optimization threshold $ \tau $ on MS-COCO dataset.}
\label{tab:ablation-mask-coco}
\end{table}

\begin{table}[ht]
\centering
\begin{tabular}{lcccc}
\toprule
\textbf{$ \tau $} & \textbf{$ \tau = 0 $} & \textbf{$ \tau = 20 $} & \textbf{$ \tau = 50 $} & \textbf{$ \tau = 100 $} \\
\midrule
mIoU $ \uparrow $ & 29.70 & 29.90 & \textbf{29.99} & 29.94 \\
\bottomrule
\end{tabular}
\caption{Ablation results on mask optimization threshold $ \tau $ on ADE20K dataset.}
\label{tab:ablation-mask-ade20k}
\end{table}

Additionally, since Unidiffuser~\cite{unidiffuser} supports both text-to-image generation and joint text-image denoising, we conduct an ablation study comparing these two methods. Specifically, we use image captions as text conditions, set the text timestep to 0, and jointly diffuse the image and mask for 1000 steps. The results in Tab.~\ref{tab:ablation-generation} show that the text2img generation method yields better downstream segmentation performance than joint denoising, reinforcing its effectiveness for our task.

\begin{table}[ht]
\centering
\begin{tabular}{lc}
\toprule
\makebox[0.2\textwidth][l]{\textbf{Generation Type}} & \makebox[0.15\textwidth][c]{\textbf{mIoU $ \uparrow $}} \\
\midrule
joint & 70.47 \\
text2img & \textbf{72.47} \\
\bottomrule
\end{tabular}
\caption{Ablation results on generation type on Psacal VOC dataset.}
\label{tab:ablation-generation}
\end{table}

Tab.\ref{tab:rebuttal-computation} reports the resource usage on ADE20K dataset using RTX 4090 GPUs. While methods like FreeMask benefit from faster sampling, they require an additional mask input. In contrast, our method jointly generates image-mask pairs in a single pass. This design sacrifices some sampling speed in exchange for better semantic alignment and simplifies training and deployment.

\begin{table}[ht]
\centering
\begin{tabular}{lc}
\toprule
\textbf{Stage} & \textbf{Time} \\
\midrule
VAE Training & $\sim$30 GPU hours \\ 
Diffusion Training & $\sim$50 GPU hours \\
Sampling (40k) & $\sim$70 GPU hours \\
% FreeMask Sampling (40k) & $\sim$40 GPU hours \\
\bottomrule
\end{tabular}
\caption{Computation time (GPU hours) on ADE20K.}
\label{tab:rebuttal-computation}
\end{table}

JoDiffusion differs from Mask2Image methods such as FreeMask in that it does not require any real semantic masks during generation. Instead, it synthesizes both the image and its pixel-level annotation from text, making it more scalable to open-domain and low-resource scenarios. In Table 3 in the paper, the performance gap between JoDiffusion and FreeMask appears small because both methods use the same amount of synthetic data and both apply filtering with pretrained segmentation models. FreeMask is already a strong baseline under this controlled setting, so large gains are not expected. To further validate the advantage of our method, we conduct an additional low-resource experiment. We compare both methods using only 2,000 inputs—either real masks (FreeMask) or GPT-generated prompts (JoDiffusion)—to synthesize 40k samples. As shown in Tab.\ref{tab:rebuttal-lowres-freemask}, our method achieves significant mIoU improvement, demonstrating stronger generalization from language-only supervision and confirming our method’s scalability without relying on annotated masks.
\begin{table}[ht]
\centering
\begin{tabular}{lcccc}
\toprule
 & \textbf{Condition} & \textbf{Data Type} & \textbf{R-50} & \textbf{Swin-T} \\
\midrule
FreeMask & 2k masks & \makecell[c]{20k real \\ 40k synthesis} & 47.50 & 48.90 \\
Ours & 2k prompts & \makecell[c]{20k real \\ 40k synthesis} & \textbf{48.11} & \textbf{50.02} \\
\bottomrule
\end{tabular}
\caption{Low-resource comparison on Mask2Former.}
\label{tab:rebuttal-lowres-freemask}
\end{table}

\subsection{Addition Visualization}

To provide more insight into our method, we present additional visualizations on Pascal VOC, MS-COCO, and ADE20K datasets.

\begin{figure}
\centering
\includegraphics[width=0.95\linewidth]{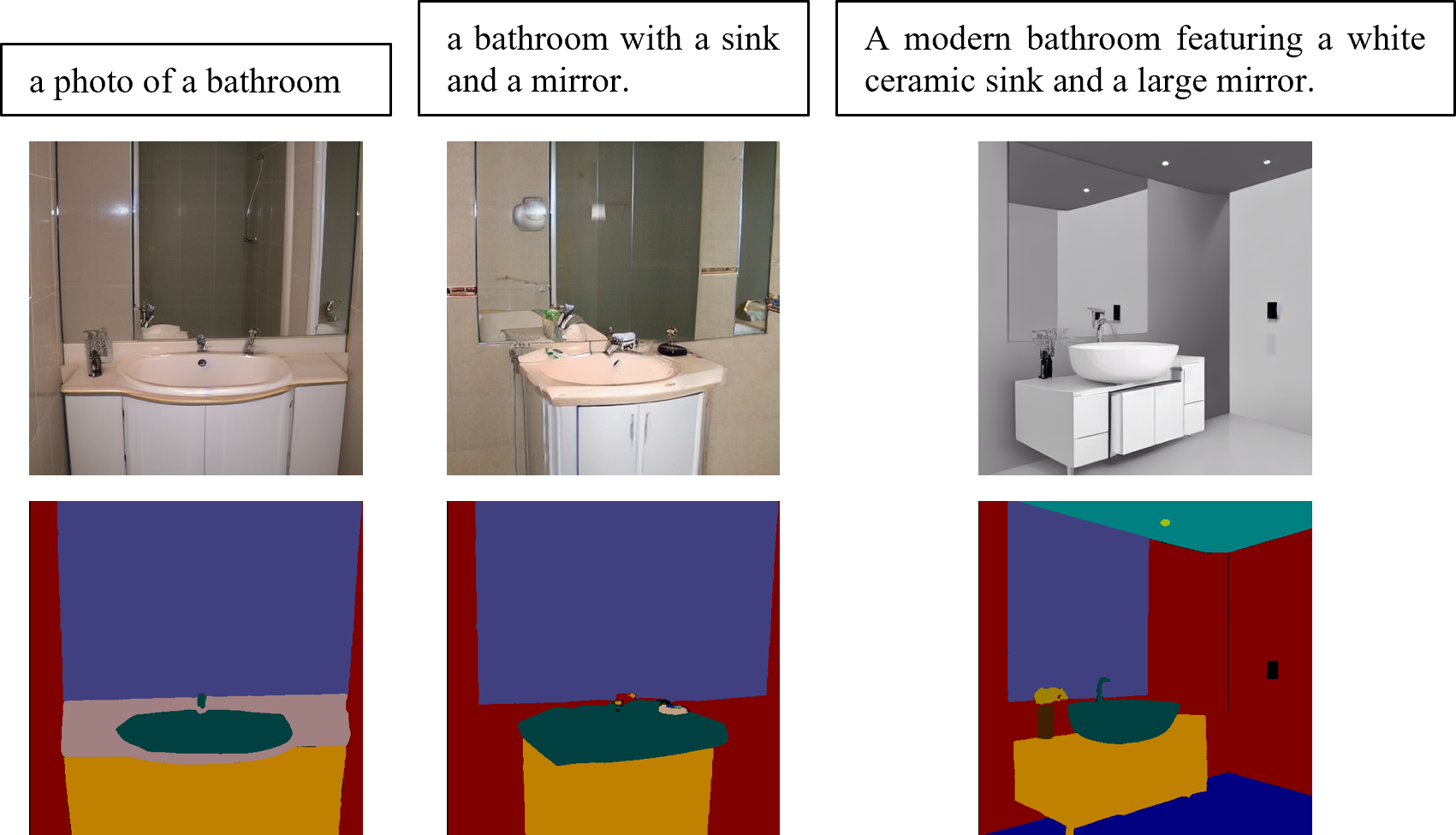}
\caption{Generation results on different text prompts.}
\label{fig:prompt}
\end{figure}

We use different prompts to generate paired results on ADE20K. As shown in Fig.~\ref{fig:prompt}, compared with the BLIP2 description in the second column, the description generated by LLM~\cite{gpt4} in the third column leads to higher quality results.

\begin{figure*}
\centering
\includegraphics[width=0.95\linewidth]{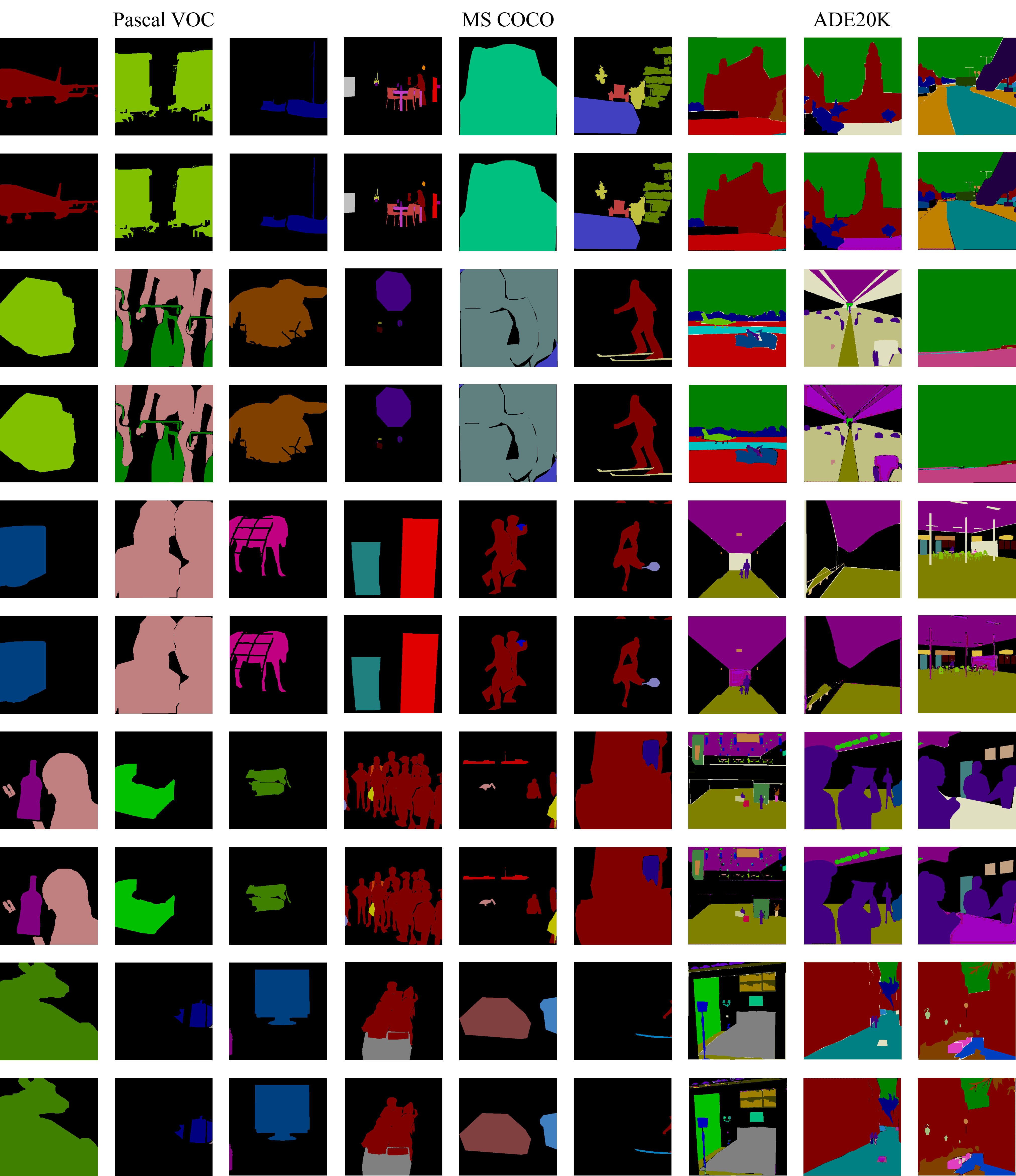}
\caption{Visualization of reconstructed pixel-level annotation masks on the validation sets of three datasets. We display the first 15 validation images of each dataset. The odd rows show the input annotation masks, while the even rows present the reconstructed results. A color map is applied for better visualization.}
\label{fig:more-vae}
\end{figure*}

Fig.~\ref{fig:more-vae} provides qualitative analysis of the Variational Autoencoder (VAE) component by visualizing reconstructed semantic segmentation masks on the validation subsets of Pascal VOC, MS-COCO, and ADE20K datasets. The results demonstrate the model's capacity to preserve fine-grained spatial details such as texture boundaries and object edges across different scene complexity levels.

\begin{figure*}
\centering
\includegraphics[width=0.95\linewidth]{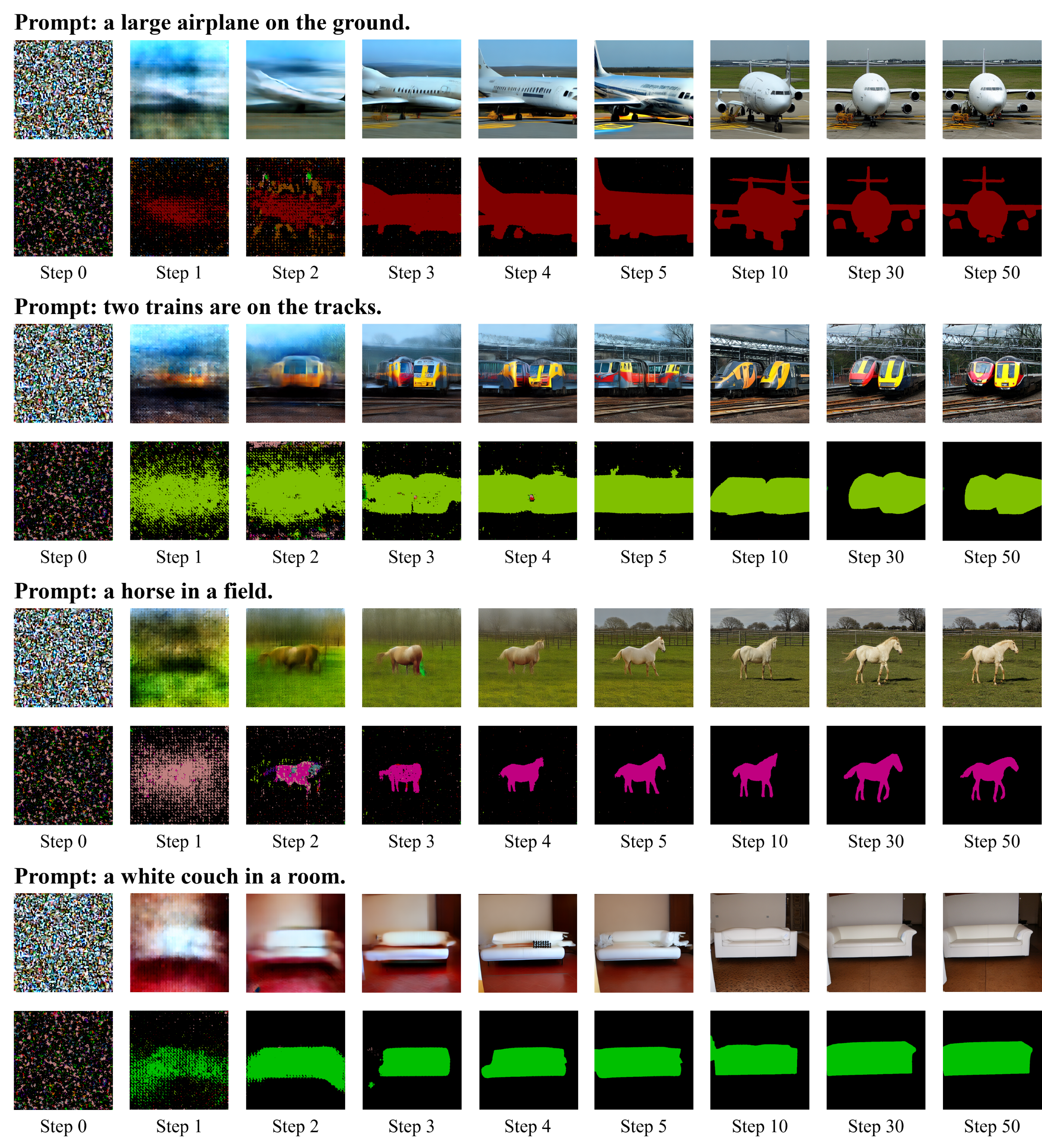}
\caption{Visualization of joint generation result at different timesteps on Pascal VOC dataset. We use validation set captions to evaluate the model's generalization ability. A color map is applied for better visualization.}
\label{fig:more-voc}
\end{figure*}

\begin{figure*}
\centering
\includegraphics[width=0.95\linewidth]{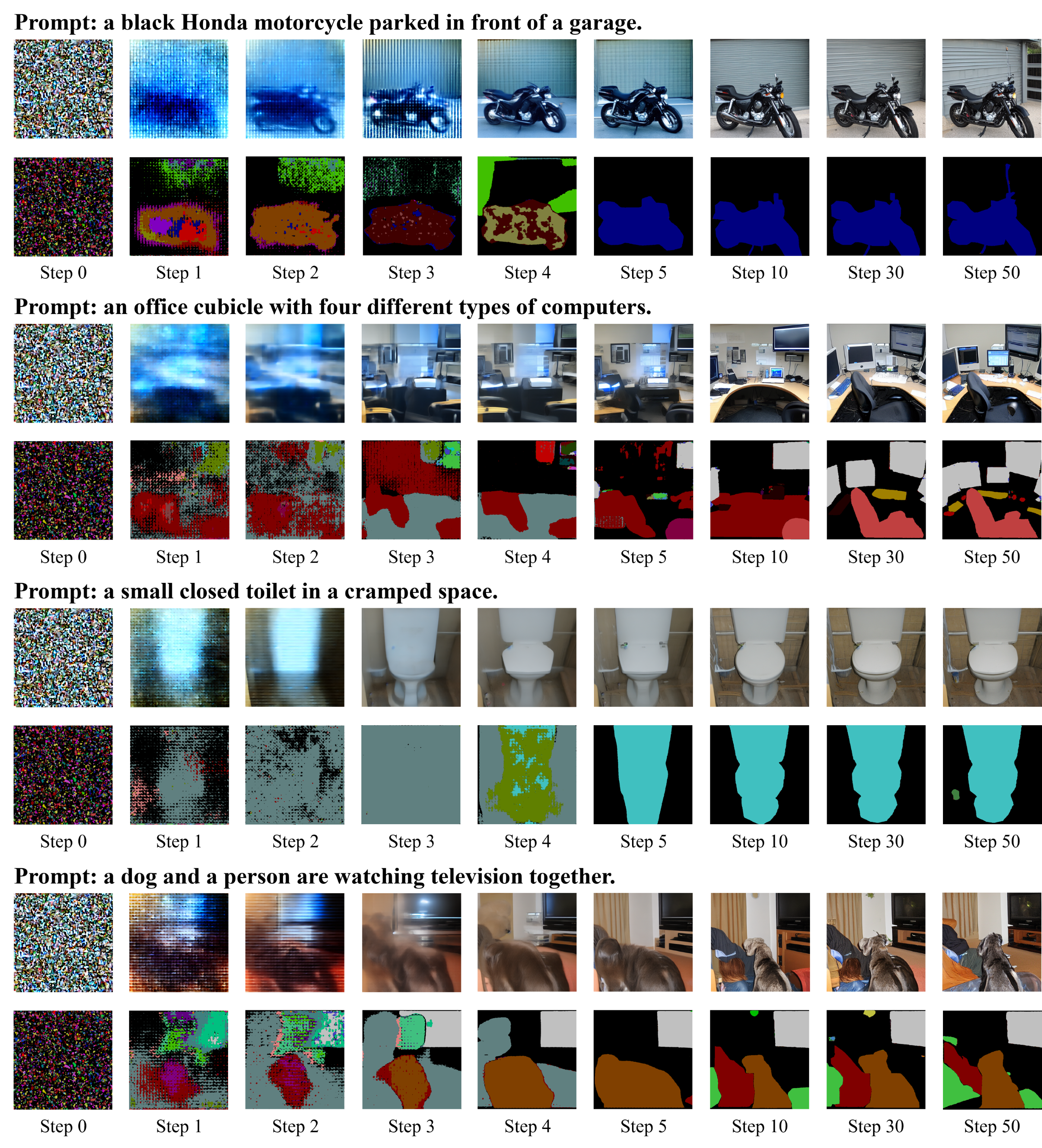}
\caption{Visualization of joint generation result at different timesteps on MS-COCO dataset. We use validation set captions to evaluate the model's generalization ability. A color map is applied for better visualization.}
\label{fig:more-coco}
\end{figure*}

\begin{figure*}
\centering
\includegraphics[width=0.95\linewidth]{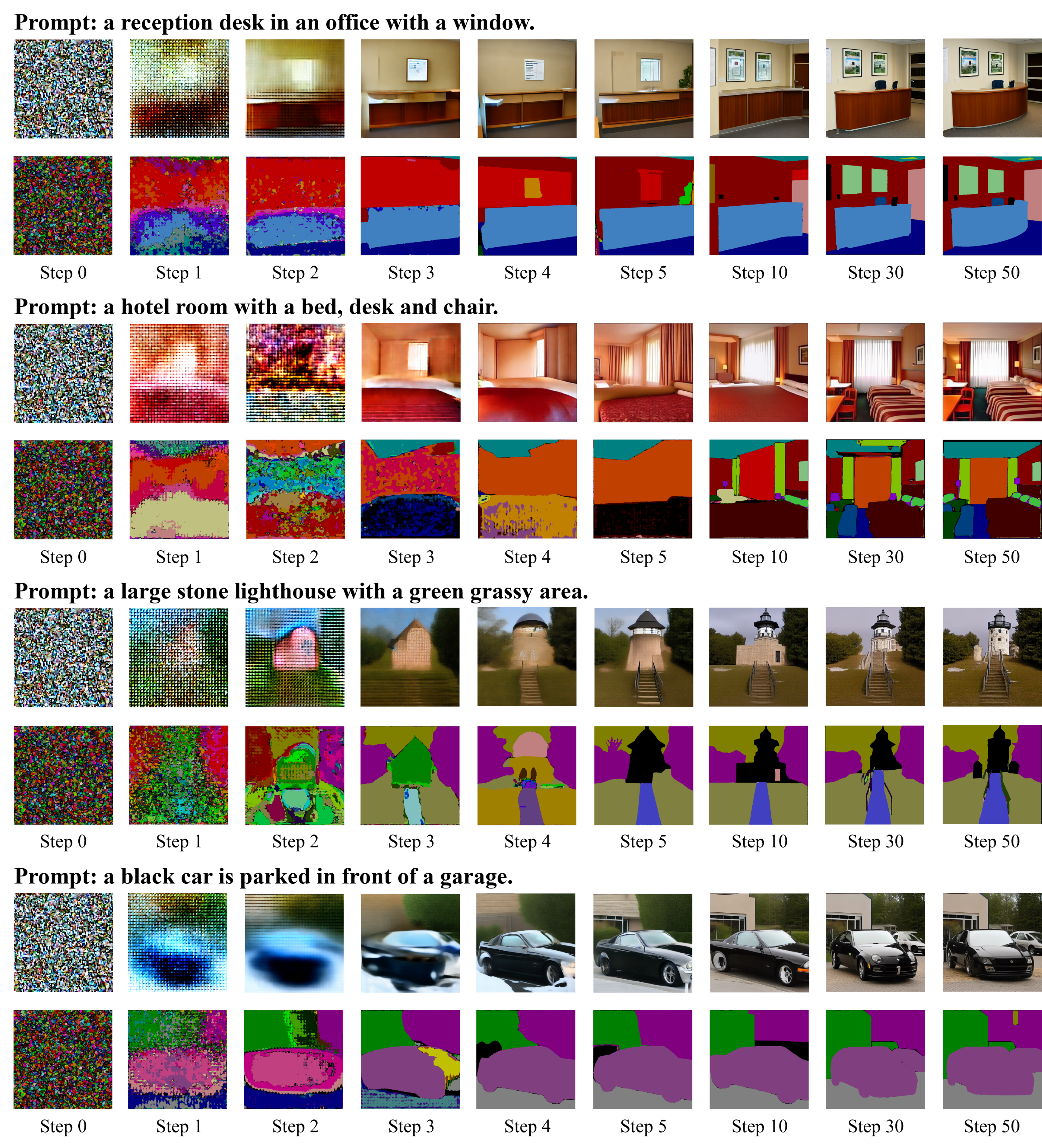}
\caption{Visualization of joint generation result at different timesteps on ADE20K dataset. We use validation set captions to evaluate the model's generalization ability. A color map is applied for better visualization.}
\label{fig:more-ade20k}
\end{figure*}

A temporal progression analysis of the joint image-mask generation process is presented in Fig.~\ref{fig:more-voc}-\ref{fig:more-ade20k}. These visualizations illustrate the iterative refinement of both RGB images and corresponding pixel-wise annotations at multiple timesteps for each dataset.

\begin{figure*}
\centering
\includegraphics[width=0.95\linewidth]{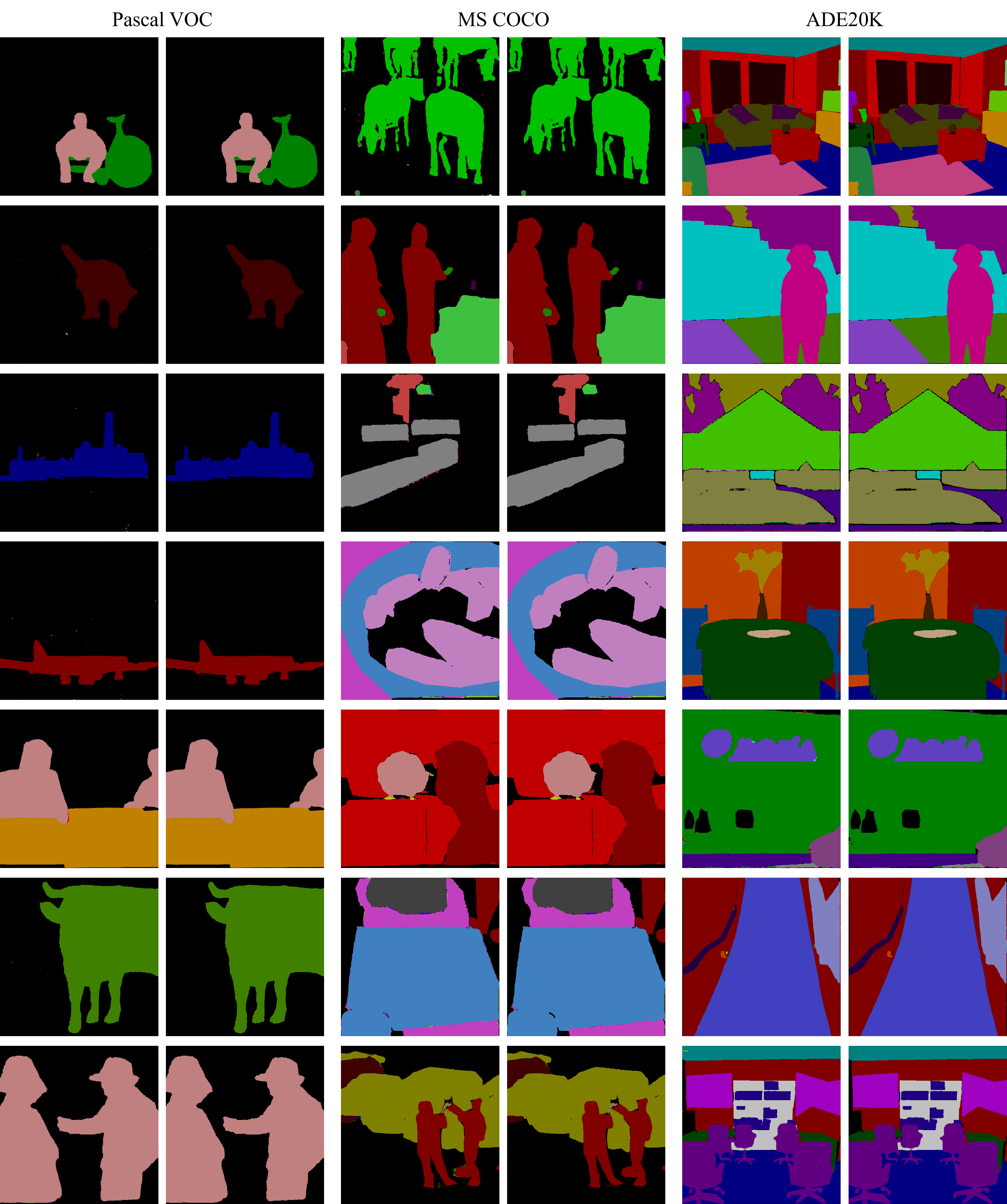}
\caption{Visualization of mask optimization strategy applied to three datasets. This refinement step reduces label inconsistencies and enhances the quality of pixel-level annotation masks. A color map is applied for better visualization. Zoom in for details.}
\label{fig:more-optim}
\end{figure*}

The efficacy of our proposed mask optimization strategy is quantitatively validated in Fig.~\ref{fig:more-optim}. This visualization compares segmentation masks before and after applying our mask optimization strategy, demonstrating improvements in boundary localization and class-specific detail preservation across all evaluated datasets.

\begin{figure*}
\centering
\includegraphics[width=0.95\linewidth]{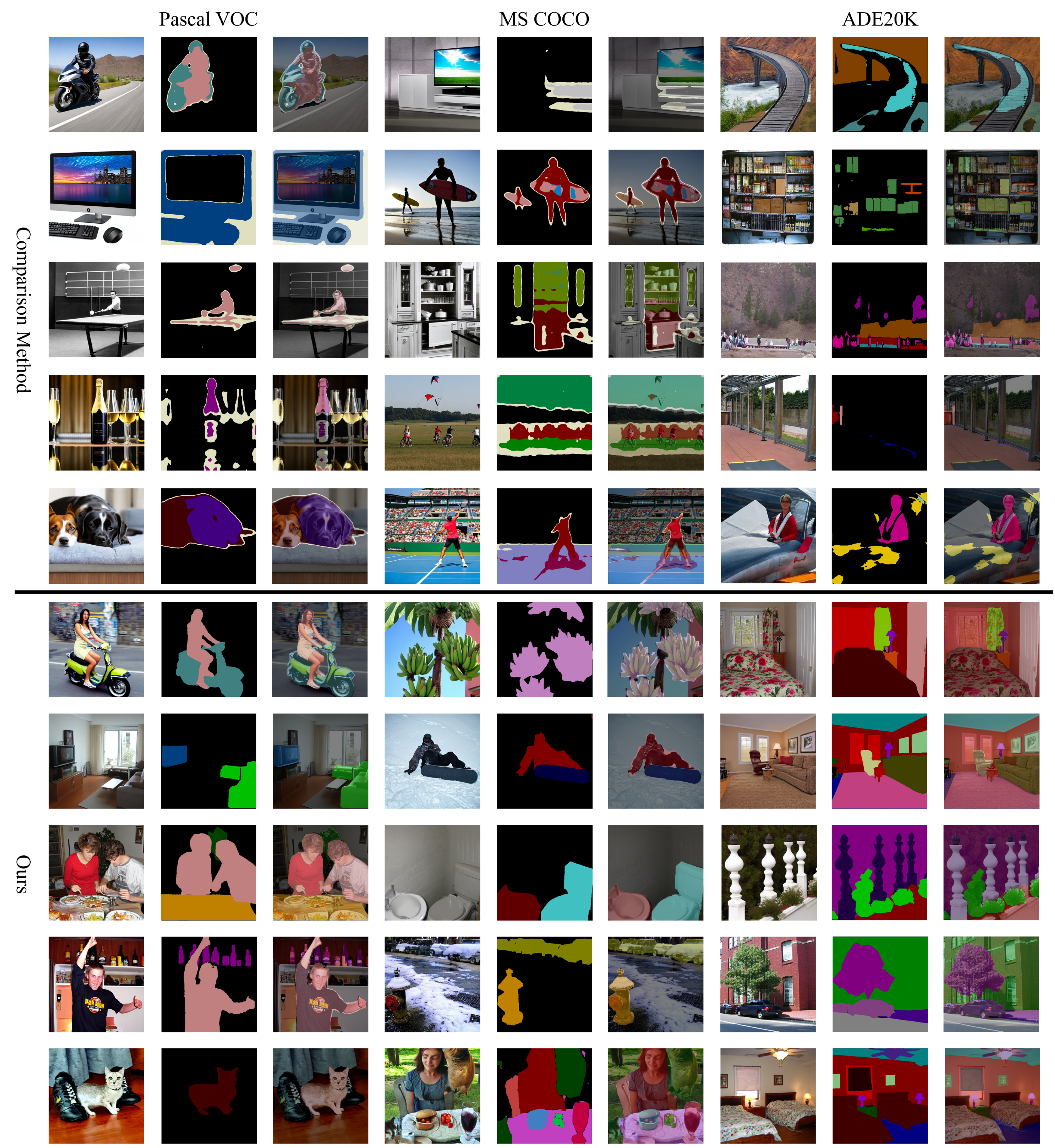}
\caption{Comparison of images and pixel-level annotation masks generated by comparsion method and our JoDiffusion framework. A color map is applied for better visualization.}
\label{fig:more-compare}
\end{figure*}

Comparative analysis in Figure~\ref{fig:more-compare} benchmarks JoDiffusion against state-of-the-art generative methods (including Dataset Diffusion and FreeMask). The results highlight JoDiffusion's superior performance in generating semantically consistent image-annotation mask pairs.

\section{More Limitation}

Despite the strong performance of our method, some limitations remain. First, while our approach can generate diverse image and pixel-level annotation mask pairs, its performance is constrained by the quality of the text captions used for generation. Second, since we rely purely on synthetic data for training segmenters in some settings, the generalization ability to real-world images may require further improvement. Future work could explore adaptive refinement techniques to enhance segmentation performance.

\clearpage

\bibliography{aaai2026}